\documentclass[10pt,a4paper]{article}
\usepackage[margin=1in]{geometry}
\usepackage{amsmath,amssymb}
\usepackage{booktabs}
\usepackage{graphicx}
\usepackage{xcolor}
\usepackage{hyperref}
\usepackage{array}
\usepackage{caption}
\usepackage{subcaption}
\usepackage{longtable}
\usepackage{float}
\usepackage{hyperref}
\hypersetup{colorlinks=true,linkcolor=blue,citecolor=blue,urlcolor=blue}

\graphicspath{{figures/}{./}}

\title{Curriculum-Driven Deep Reinforcement Learning for Autonomous Far-Range Rocket Landing}
\author{
\begin{tabular}{cc}
\begin{tabular}{c}
Atin Chowdhury\\[3pt]
\href{mailto:atin.c23@iiits.in}{atin.c23@iiits.in}
\end{tabular}
&
\begin{tabular}{c}
Dr. Piyush Joshi\\[3pt]
\href{mailto:piyush.j@iiits.in}{piyush.j@iiits.in}
\end{tabular}
\end{tabular}
\\[1em]
\small Department of Computer Science and Engineering\\
\small Indian Institute of Information Technology, Sri City, Andhra Pradesh, India
}
\date{}

\begin{document}
\maketitle

\begin{abstract}
Autonomous vertical landing of reusable launch vehicles is a critical yet formidable challenge in aerospace engineering, primarily due to highly nonlinear dynamics, stringent precision requirements, and inherent environmental uncertainties. While reinforcement learning (RL) has emerged as a promising alternative to traditional model-based control, existing approaches frequently rely on flat training paradigms and highly simplified initial conditions, drastically limiting their real-world applicability. This paper proposes a comprehensive curriculum-driven deep reinforcement learning framework designed to master both far-range navigation and precision landing. By employing a three-stage curriculum augmented with a critic warm-up procedure and a rehearsal mechanism to prevent catastrophic forgetting, alongside a high-dimensional observation vector utilizing a full $3\times3$ rotation-matrix, the agent is trained to navigate from an initial horizontal distance of $1{,}000$ to $1{,}500$ units vastly farther than prior benchmarks before executing a soft touchdown. Crucially, we introduce a redesigned reward function that reduces unnecessary computational overhead and addresses limitations of earlier formulations, including poorly calibrated reward components and potentially conflicting optimization objectives. The redesigned function bounds the per-step reward magnitude by (O(1)), while its progress term accumulates the total distance closed over the trajectory. The available $100{,}000$-episode results report a 99.98\% success rate for the proposed policy and 68.31\% for the baseline under the evaluation conditions. In the evaluation, the proposed policy retained 69.1\% of its initial fuel capacity and achieved a mean final distance from the pad centre of 7.07 units, compared with 13.10 units for the baseline. Landing visualizations are available in \href{https://drive.google.com/file/d/1JeAk50SnNZV-DMDPuOA74ZvhBRDYV8xo/view?usp=drive_link}{Video 1} and \href{https://drive.google.com/file/d/1KTEet75bhmqcttfkGxuTcoHXAw8AlLG_/view?usp=drive_link}{Video 2}.
\end{abstract}
\section{Introduction}

Vertical landing of a reusable rocket stage is one of the most demanding problems in aerospace control. Systems in operational use, such as the Falcon~9 first stage, achieve high reliability through carefully engineered model-based guidance, sensor fusion, and real-time trajectory optimisation~\cite{blackmore2016,elke2021}. The strength of these designs is also their limitation: they rely on accurate dynamical models and on extensive manual calibration, and their applicability becomes questionable in unknown, uncertain, or extraterrestrial environments where such models and calibration are unavailable~\cite{ross2022}. Reinforcement learning (RL) offers an alternative, since a policy can in principle be discovered by direct interaction with a simulator without a hand-designed control law.

Prior work has explored RL for rocket landing, utilizing techniques such as PPO, curriculum learning, and imitation learning. However, many of these studies operate on simplified dynamics or assume initial conditions where the vehicle is already in close proximity to the landing pad, making the lateral approach a trivial component of the problem. A notable recent baseline by Chowdhury et al.~\cite{chowdhury2024icaart} reported 84\% success over 100 episodes, with the rocket starting close to the pad. 


The proposed framework addresses these limitations by forcing the agent to master a far-range approach from $1{,}000$ to $1{,}500$ units away. We introduce a three-stage curriculum with explicit weight transfer, a novel critic warm-up procedure, and a rehearsal mechanism to prevent catastrophic forgetting. A high-dimensional rotation-matrix observation replaces the quaternion plus proxy combination and includes body-frame projections of the horizontal position and velocity so that the network is directly told which tilt key will move the nose against the drift. The reward function is redesigned into a compact form with $O(1)$ per-step magnitude that explicitly rewards long-range navigation and braking alignment. The evaluation reports a $99.98\%$ success rate for the proposed policy under the evaluation conditions described in Section~\ref{sec:results}.

The contributions of this paper are the following. First, a three-stage curriculum (far approach, soft descent, full landing) is designed and evaluated, together with a critic warm-up procedure and a rehearsal mechanism. Second, a high-dimensional rotation-matrix observation with body-frame velocity projections is introduced and used with a compact policy network. Third, a reward function is redesigned term by term, reducing the number of hand-tuned constants from fifteen to eight and rendering the per-step magnitude $O(1)$ instead of $O(10^2)$. Fourth, the initial horizontal reach is widened to $1{,}000$--$1{,}500$ units relative to prior work. Fifth, the training dynamics of the three curriculum stages are analysed through reward, episode-length, and critic-conditioning curves. Sixth, a $100{,}000$-episode evaluation is reported in Section~\ref{sec:results}.

\section{Related Work}

The application of reinforcement learning to autonomous guidance and control has grown rapidly over the past decade. Early efforts focused on simplified 2D environments, such as the LunarLander benchmark from Gymnasium, which served to demonstrate that agents could learn basic descent policies through trial and error~\cite{sadavarte2021}. However, these 2D testbeds fundamentally omit the coupled three-dimensional rotational dynamics that are central to real rocket landing.

On the classical side, lossless convexification and successive convexification have been the workhorse approaches for powered descent guidance~\cite{acikmese2013lossless,blackmore2010minimum}. These methods rely on accurate dynamical models and solve a convex optimisation problem at every guidance cycle, which is computationally expensive and requires careful hand-tuning of the trust-region and discretisation parameters. More recently, Gaudet, Linares and Furfaro have explored deep reinforcement learning for powered descent guidance in a series of papers~\cite{gaudet2020deep,gaudet2020terminal,gaudet2021mars}, demonstrating that RL can match or exceed convex optimisation in accuracy while being several orders of magnitude faster at inference time.

In the domain of 3D rocket landing, various strategies have been employed to handle the high-dimensional state space and sparse rewards. Rathod et al.~\cite{rathod2023} trained PPO inside a Unity-based Starship-like environment, utilizing curriculum learning and imitation learning to achieve near-perfect success while discovering fuel-efficient suicide-burn manoeuvres. To address the exploration challenge inherent in landing tasks, Jiang et al.~\cite{jiang2024} proposed a Random-Annealing Jump-Start framework, which couples classical feedback control with RL to bootstrap the learning process and accelerate convergence. Su et al.~\cite{su2022} proposed a continuous-control DRL agent with multi-stage reward shaping, focusing on real-time fuel efficiency. All three of these works rely, to varying degrees, on a hand-tuned controller or synthetic curriculum scaffold to bootstrap the learning; our curriculum is fully learning-driven and does not require a classical controller at any stage.

More recent studies have explored advanced algorithmic and architectural variations. Zhu et al.~\cite{zhu2025adaptppo} introduced AdaptPPO, an adaptive variant of Proximal Policy Optimization designed for precision rocket landing, demonstrating improved sample efficiency. Zhang et al.~\cite{ieee2022thrust} developed a 3-DoF thrust-vectored rocket model and applied PPO to learn integrated guidance and control. For model-based approaches, Rossi et al.~\cite{ieee2026robust} introduced an adaptive weighting mechanism to stabilize multiobjective learning and achieved an 82.1\% success rate under randomized conditions, outperforming standard MBRL baselines. Propulsive landing of launchers' first stages was investigated using a 6-DOF guidance approach with deep neural network policies by Dubois et al.~\cite{sciencedirect2024propulsive}, emphasizing reward shaping for fuel-optimal landings. To resolve the trade-off between landing accuracy and fuel consumption, Kim et al.~\cite{ieee2025hierarchical} proposed a hierarchical architecture combining RL-based guidance with a Sliding Mode Controller tracker, demonstrating improved accuracy over 500 Monte Carlo simulations. Furthermore, hardware-software co-design has been explored to achieve real-time inference, such as the AHA-TD3 framework in~\cite{nature2026fpga}, which reported significant reductions in inference latency using FPGA acceleration.

Curriculum learning has itself been formalised and surveyed extensively~\cite{narvekar2020curriculum}. Narvekar et al. identify several dimensions along which curricula differ: task decomposition, task sequencing, and transfer mechanism. Our curriculum combines a fixed hand-designed sequence (far approach, soft descent, full landing) with two transfer mechanisms critic warm-up and rehearsal mixing that are, to our knowledge, novel in the context of rocket landing. Prior work on rehearsal or experience replay in RL~\cite{schaul2015prioritized} has shown that mixing past and current experience stabilises learning and prevents forgetting; we apply the same principle at the level of curriculum stages.

The most closely related recent work to the proposed framework is the study by Chowdhury et al.~\cite{chowdhury2024icaart}. That paper introduced a custom three-dimensional physics simulator incorporating gravity, thrust, off-centre torque, angular damping, and finite fuel, and it provided a direct comparison between PPO, DQN, and Double DQN. PPO was reported at 84\% success over 100 episodes, while the value-based methods failed to learn effectively. While this establishes a strong foundation for 3D RL landing, it has notable limitations that the proposed framework specifically addresses. The training in that baseline is entirely flat, forcing the agent to discover attitude, braking, and lateral control simultaneously. The success criterion is coarse, and the evaluation is limited to only 100 episodes. Most importantly for the physical realism of the task, the initial horizontal displacement in that study is very small, meaning the rocket begins almost directly over the pad centre.

The present work differs from these prior studies in several key aspects. We employ a multi-stage curriculum that transfers weights between distinct sub-tasks, coupled with a novel critic warm-up phase and a rehearsal mechanism. We adopt a full $3\times 3$ rotation matrix observation to exploit rotational symmetry, drastically reducing the network size. Furthermore, we vastly expand the initial horizontal distribution to $1{,}000$--$1{,}500$ units to force a genuine, multi-phase lateral manoeuvre from far away. Finally, the evaluation contains $100{,}000$ episodes, providing a larger evaluation sample than the 100-episode evaluation reported for the baseline.

\section{Environment and Problem Formulation}

The simulator used in this work is a lightweight, deterministic, per-frame physics integrator running at $60$~Hz, with three physics frames accumulated into one decision of the reinforcement-learning agent. Throughout this paper, unless otherwise noted, lengths are measured in simulator units, times in seconds, and ``decision'' refers to one agent step (three physics frames). The simulation constants are inherited from the baseline environment: gravitational acceleration is $-0.007$ units/frame$^2$, thrust force is $0.02$ units/frame$^2$, the torque coupling factor is $250$, angular damping is $0.985$, thrust consumes $12$ units of fuel per second, tilt consumes $2.5$ units of fuel per second, the tilt limit is $80^\circ$, the tilt ramp per frame is $1.5^\circ$, the landing altitude is $13$ units, the pad radius $R_{\text{pad}} = 16$ units, the vertical speed limit for a successful landing is $-0.15$ units/frame, the horizontal speed limit for a successful landing is $0.25$ units/frame, the workspace ceiling is $2500$ units with a horizontal workspace limit of $3000$ units. The fuel capacity in this work is $F_{\max} = 1000$ units, double the $500$ units of the baseline, to provide sufficient margin for the long-distance navigation.

For touchdown-quality assessment, we use four conditions: (i) tilt angle $\theta \leq 10^\circ$; (ii) horizontal distance from the pad centre $d_h \leq R_{\text{pad}}=16$ units; (iii) vertical velocity $v_z \geq -0.15$ units/frame; and (iv) horizontal speed $v_h \leq 0.25$ units/frame. A touchdown satisfies the joint four-condition criterion only when all four conditions hold simultaneously. The tilt condition is assessed from the measured touchdown attitude.

The ship's attitude is represented by a full orthonormal $3\times 3$ rotation matrix whose columns are the ship's own axes expressed in world coordinates. Writing the columns as the ship body $X$ axis, the nose direction, and the ship body $Z$ axis respectively, the rotation update at every frame is
\begin{equation}
R \leftarrow R \cdot R_x(\omega_P) \cdot R_z(\omega_R),
\end{equation}
where $\omega_P$ and $\omega_R$ are the pitch and roll angular rates in degrees per frame and $R_x$ and $R_z$ are rotations about the ship's own $X$ and $Z$ axes. When thrust is on, the acceleration is applied along the nose direction and a torque proportional to the commanded tilt is added to the angular rates.

The observation vector is $29$-dimensional. Using zero-based indexing, its components are
\begin{equation}
\begin{aligned}
\sigma_t = \bigl[\,& x_z,\; n_z,\; z_z,\; \tilde{\omega}_P,\; \tilde{\omega}_R,\; \tilde{\theta}_x,\; \tilde{\theta}_z,\\
& \tilde{v}_z,\; \tilde{h},\; F/F_{\max},\; \tilde{m},\; \tilde{x},\; \tilde{y},\; \tilde{v}_x,\; \tilde{v}_y,\\
& \tilde{x}^{(x)}_{\text{body}},\; \tilde{x}^{(z)}_{\text{body}},\; \tilde{v}^{(x)}_{\text{body}},\; \tilde{v}^{(z)}_{\text{body}},\; n_x,\; n_y,\\
& \tilde{z}_{\text{err}},\; \tilde{d},\; \tilde{v}_{\text{rad}},\; \tilde{m}_h,\; \tilde{x}_{\text{fine}},\; \tilde{y}_{\text{fine}},\; \tilde{v}_{x,\text{fine}},\; \tilde{v}_{y,\text{fine}} \,\bigr],
\end{aligned}
\end{equation}
where the tilde denotes a fixed scaling. Components 0--2 are the vertical components of the ship body $X$ axis, the nose, and the ship body $Z$ axis respectively. Components 15--18 are the horizontal position and velocity projected onto the ship body $X$ and $Z$ axes, and they explicitly tell the agent which tilt key opposes the current drift; note that components 15 and 16 are position projections ($\tilde{x}^{(x)}_{\text{body}}$, $\tilde{x}^{(z)}_{\text{body}}$) and components 17 and 18 are velocity projections ($\tilde{v}^{(x)}_{\text{body}}$, $\tilde{v}^{(z)}_{\text{body}}$). Components 21--24 provide the height error to the hover point, horizontal distance, radial velocity toward the pad, and horizontal braking margin. Components 25--28 provide fine-scale copies of position and velocity near the pad. The stopping margin $\tilde{m}$ in component 10 uses the net deceleration $a_{\text{net}} = 0.013$ units/frame$^2$, which is the thrust force minus the magnitude of gravity ($0.020 - 0.007$), i.e.\ the maximum possible vertical deceleration available to the agent, and hence the physically correct constant for the ``distance-to-stop'' calculation.

The action is a joint discrete tuple
\begin{equation}
a_t = (a_{\text{thr}},\, a_{\text{pitch}},\, a_{\text{roll}}) \in \{0,1\} \times \{0,1,2\} \times \{0,1,2\},
\end{equation}
giving $18$ joint choices. The thrust channel is binary, the pitch channel selects among no tilt, tilt forward, and tilt backward, and the roll channel selects among no tilt, tilt left, and tilt right. Every channel is chosen simultaneously.

\section{Reward Function Design}

The reward function serves as the critical communication channel between the physical problem and the learning algorithm. A well-designed reward function must guide the agent toward the desired behavior without introducing unnecessary computational complexity or conflicting objectives. In this section, we detail the reward structure for each stage of the curriculum. We then provide a comprehensive comparison against the reward function used in the baseline study, highlighting the improvements in both efficiency and interpretability.

\subsection{Stage One Reward (Far Approach)}

Stage one is trained to navigate from a far distance ($1{,}000$--$1{,}500$ units) to a hover point above the pad. The reward at each decision combines progress, velocity caps, and alignment incentives. We define the horizontal distance to the pad as $d_h$, the horizontal speed as $v_h$, and the height error as $\Delta z = z - z_{\text{hov}}$, with $z_{\text{hov}} = 300$. The reward is
\begin{equation}
\begin{aligned}
r^{(1)}_t = {}& w_{\text{prog}} (d_{h,t-1} - d_{h,t}) + w_{\text{prog\_z}} (|\Delta z_{t-1}| - |\Delta z_t|)\\
& - w_{\text{ov}} \min\!\left(\frac{\max(0, v_h - v_{\text{cap}})}{0.3}, 2\right) - w_{\text{ovz}} \min\!\left(\frac{\max(0, v_{\text{des}} - v_z)}{|v_{\text{des}}| + 0.2}, 2\right)\\
& - w_{\text{orient}} \left(\frac{\theta}{90^\circ}\right)^2 - w_\omega\,\omega - w_{\text{vh}} \min(v_h, 1) \exp\!\left(-\frac{d_h}{80}\right)\\
& - w_{\text{time}} + r_{\text{align}} + 0.3\,\mathbb{1}_{\text{in\_zone}},
\end{aligned}
\end{equation}
where $v_{\text{cap}} = \sqrt{2 A_H d_h} + 0.05$ is the horizontal braking speed profile, and $v_{\text{des}}$ is the vertical braking profile. The term $r_{\text{align}}$ shapes the thrust direction:
\begin{equation}
r_{\text{align}} = w_{\text{accel}} a_{\text{thr}} n_h \max(\hat{n} \cdot \hat{d}, 0) \mathbb{1}_{\text{accel}} + w_{\text{brake}} a_{\text{thr}} n_h \max(\hat{n} \cdot (-\hat{v}_h), 0) \mathbb{1}_{\text{brake}},
\end{equation}
where $n_h = \sqrt{n_x^2 + n_y^2}$ is the horizontal component of the nose. This encourages the agent to tilt toward the pad when accelerating and counter-tilt against its velocity when braking. An episode terminates successfully as soon as the agent satisfies the hover condition (horizontal distance $\le 10$ units, height error $\le 15$ units, horizontal speed $\le 0.06$, vertical speed $\le 0.10$, tilt $\le 8^\circ$) for $15$ consecutive decisions, or terminates with failure at the workspace bounds.

\subsection{Stage Two and Stage Three Reward (Soft Descent and Full Task)}

Stages two and three share a landing reward, but with different initial distributions and tilt-penalty weights. Stage two starts near the pad and learns vertical braking; stage three integrates far-range navigation with landing. The reward is anchored on a soft-descent reference trajectory
\begin{equation}
v_{\text{des}}(h) = -\max\!\left(V_{\text{soft}},\; \sqrt{2\,A_{\text{des}}\,h}\right),
\end{equation}
with $V_{\text{soft}} = 0.07$ and $A_{\text{des}} = 0.0078$, and $h = \max(0,\, z - z_{\text{pad}})$. The landing reward at each decision is
\begin{equation}
\begin{aligned}
r^{(2,3)}_t = {}& -w_s \min\!\left(\frac{\max(0,\, v_{\text{des}} - v_z)}{|v_{\text{des}}| + 0.2},\, 2\right) - w_{\text{slow}} \min\!\bigl(\max(v_z + 0.05,\, 0),\, 1\bigr)\\
& - w_o \left(\frac{\theta}{90^\circ}\right)^2 - w_{o,\text{low}} \exp\!\left(-\frac{h}{80}\right) \min\!\left(\frac{\theta}{10^\circ}, 3\right)\\
& - w_\omega\,\omega - w_{p,\text{low}} \exp\!\left(-\frac{h}{100}\right) \min\!\left(\frac{d_h}{16}, 3\right)\\
& - w_{vh} \min(v_h, 1) \exp\!\left(-\frac{d_h}{80}\right) - w_{\text{ov}} \min\!\left(\frac{\max(0, v_h - v_{\text{cap}})}{0.3}, 2\right)\\
& - w_t\, a_{\text{thr}} - w_\tau + w_{\text{prog}} (d_{h,t-1} - d_{h,t}) + r_{\text{align}},
\end{aligned}
\end{equation}
with the weights listed in Table~\ref{tab:rw_weights}. The term $w_{\text{prog}} (d_{h,t-1} - d_{h,t})$ is positive whenever the agent reduces its distance to the pad and, unlike the per-step penalties, it accumulates over the episode. For a stage-three episode beginning at $d_h \approx 1131$ and ending near $7$, the total accumulated progress reward is approximately $w_{\text{prog}} \cdot 1124 \approx 90$ raw units (before scaling by $0.1$), i.e.\ $+9$ on the scaled return. This is the mechanism that drives long-range navigation, not the pad-distance penalty, which is deliberately saturated at large distances. The pad-distance penalty and horizontal-velocity penalty serve only as fine-grained shaping near the pad, and they are active only in the last few dozen units of the traverse. In stage three, the tilt penalty weights $w_o$ and $w_{o,\text{low}}$ are set to zero to avoid training a separate upright-recovery objective; the agent uses tilt purely for navigation and braking.

\begin{table}[H]
\centering
\small
\caption{Landing reward weights used in stages two and three. Stage two includes tilt penalties to enforce uprightness, while stage three disables them ($w_o = w_{o,\text{low}} = 0$) to focus on composite navigation.}
\label{tab:rw_weights}
\begin{tabular}{llc}
\toprule
Symbol & Meaning & Value \\
\midrule
$w_s$ & soft-descent reference tracking & $0.4$ \\
$w_{\text{slow}}$ & near-ground slow-speed shaping & $0.5$ \\
$w_o$ & tilt penalty (stage two / stage three) & $0.3$ / $0.0$ \\
$w_{o,\text{low}}$ & near-ground tilt penalty (stage two / stage three) & $0.3$ / $0.0$ \\
$w_\omega$ & angular-rate penalty & $0.05$ \\
$w_{p,\text{low}}$ & near-ground pad-distance penalty & $0.1$ \\
$w_{vh}$ & horizontal-velocity penalty & $0.2$ \\
$w_{\text{ov}}$ & horizontal overspeed penalty & $0.6$ \\
$w_t$ & thrust-cost penalty & $0.002$ \\
$w_\tau$ & per-decision time penalty & $0.015$ \\
$w_{\text{prog}}$ & horizontal progress reward & $0.08$ \\
$w_{\text{accel}}$ & thrust alignment for acceleration & $0.45$ \\
$w_{\text{brake}}$ & thrust alignment for braking & $0.90$ \\
\midrule
$V_{\text{soft}}$ & target touchdown speed & $0.07$ \\
$A_{\text{des}}$ & design deceleration & $0.0078$ \\
$R_{\text{pad}}$ & pad radius & $16$ \\
\bottomrule
\end{tabular}
\end{table}

\subsection{Terminal Adjustments}

Upon the termination of an episode, specific terminal rewards or penalties are applied to reinforce the desired final outcomes. On touchdown, the softness $\eta = 1 - \min(|v_z|/V_{z,\text{lim}},\, 1)$ is computed with $V_{z,\text{lim}} = 0.15$, and a successful landing receives
\begin{equation}
r_{\text{success}} = 40 + 10\,\eta + 5\left(1 - \frac{\theta}{10^\circ}\right),
\end{equation}
while a crash receives
\begin{equation}
\begin{aligned}
r_{\text{crash}} = -\biggl[&15 + 15\min\!\left(\frac{\max(-v_z + 0.15,\, 0)}{2}, 1\right) + 5\min\!\left(\frac{\max(\theta - 10^\circ,\, 0)}{45^\circ}, 1\right)\\
&+ 8\min\!\left(\frac{\max(d_h - 16,\, 0)}{100}, 1\right) + 8\min\!\left(\frac{\max(v_h - 0.25,\, 0)}{1}, 1\right)\biggr].
\end{aligned}
\end{equation}
Leaving the workspace applies a penalty of $-30$. The per-decision reward and the terminal adjustment together are multiplied by a global scale of $0.1$. Because the progress term accumulates positively over the episode, and the per-decision penalties are bounded below by approximately $w_\tau + w_s + w_{\text{slow}} \approx 0.915$, the total return for a successful stage-three episode is of order $+10$, consistent with the reward curves reported in Section~\ref{sec:diagnostics}. For an unsuccessful episode, the progress term is smaller (the agent may not have closed the full distance) and the terminal crash penalty dominates, so the return is strongly negative.

\subsection{Comparison with the Baseline Reward}

The baseline reward function~\cite{chowdhury2024icaart} consists of eight shaping terms with fifteen hand-tuned constants, plus a compound terminal reward, clipped to $[-500, +500]$ at every step. Because the per-step reward magnitude is $O(10^2)$ and the clipping bound is aggressive, the critic's gradient magnitude dominates the shared actor-critic gradient update early in training. This effectively freezes the actor and prevents policy improvement until the critic's error has substantially decreased. Table~\ref{tab:reward_compare} gives a term-by-term comparison.

\begin{table}[H]
\centering
\small
\caption{Term-by-term comparison between the reward of Chowdhury et al.~\cite{chowdhury2024icaart} and the present reward.}
\label{tab:reward_compare}
\begin{tabular}{p{2.9cm}p{5.2cm}p{5.4cm}}
\toprule
Concept & Baseline of~\cite{chowdhury2024icaart} & Present work \\
\midrule
Number of terms & eight shaping plus one compound terminal term & eight shaping plus three explicit terminal terms \\
\midrule
Hand-tuned constants & fifteen & eight \\
\midrule
Per-step clipping & interval $[-500,\,+500]$ & none, but per-episode return scaled by $0.1$ \\
\midrule
Reward magnitude & $O(10^2)$ per step & $O(1)$ per step \\
\midrule
Horizontal progress term & additive term with weight $0.5$ on normalised change of horizontal distance & explicit progress reward $w_{\text{prog}} (d_{t-1} - d_t)$, combined with directional alignment shaping for thrust \\
\midrule
Angular-rate penalty & based on $\sin$ of decoupled pitch and yaw angles & true body-frame norm $\sqrt{\omega_P^2 + \omega_R^2}$ \\
\midrule
Motion penalty & coupled velocity term with coefficient $C_{v_z} = 7.5$ & horizontal speed penalised separately from vertical speed, with distance-dependent scaling \\
\midrule
Alignment bonus & step-wise bonus computed from pitch and yaw angles & concave penalty on true tilt, plus explicit thrust alignment incentives for acceleration and braking \\
\midrule
Action-dependent feedback & thrust-saturation exponential and tilt-direction reward & simple thrust cost $0.002$; direction learning left to PPO \\
\midrule
Velocity-based shaping & banded reward with three thresholds & smooth soft-descent reference $v_{\text{des}}(h)$ and horizontal braking speed cap $v_{\text{cap}}(d_h)$ \\
\midrule
Distance penalty & clipped linear combination of altitude and horizontal offsets & direct saturated pad-distance penalty, amplified near the ground \\
\midrule
Compound distance penalty & hyperbolic tangent of Euclidean offset & redundant with the pad-distance and horizontal-velocity terms; not used \\
\bottomrule
\end{tabular}
\end{table}

The reward magnitude is $O(1)$ in the present work versus $O(10^2)$ in the baseline, so per-episode returns stay within $\pm 15$ instead of $O(10^4)$ and actor and critic gradients remain balanced. The soft-descent and horizontal braking references cannot be gamed by moving faster than the ship can brake. The combined effect is visible in the training curves of Section~\ref{sec:diagnostics}, where value loss declines smoothly and policy improvement begins immediately after critic warm-up.

\section{Proposed Architecture}

\begin{figure}[H]
\centering
\includegraphics[width=\textwidth]{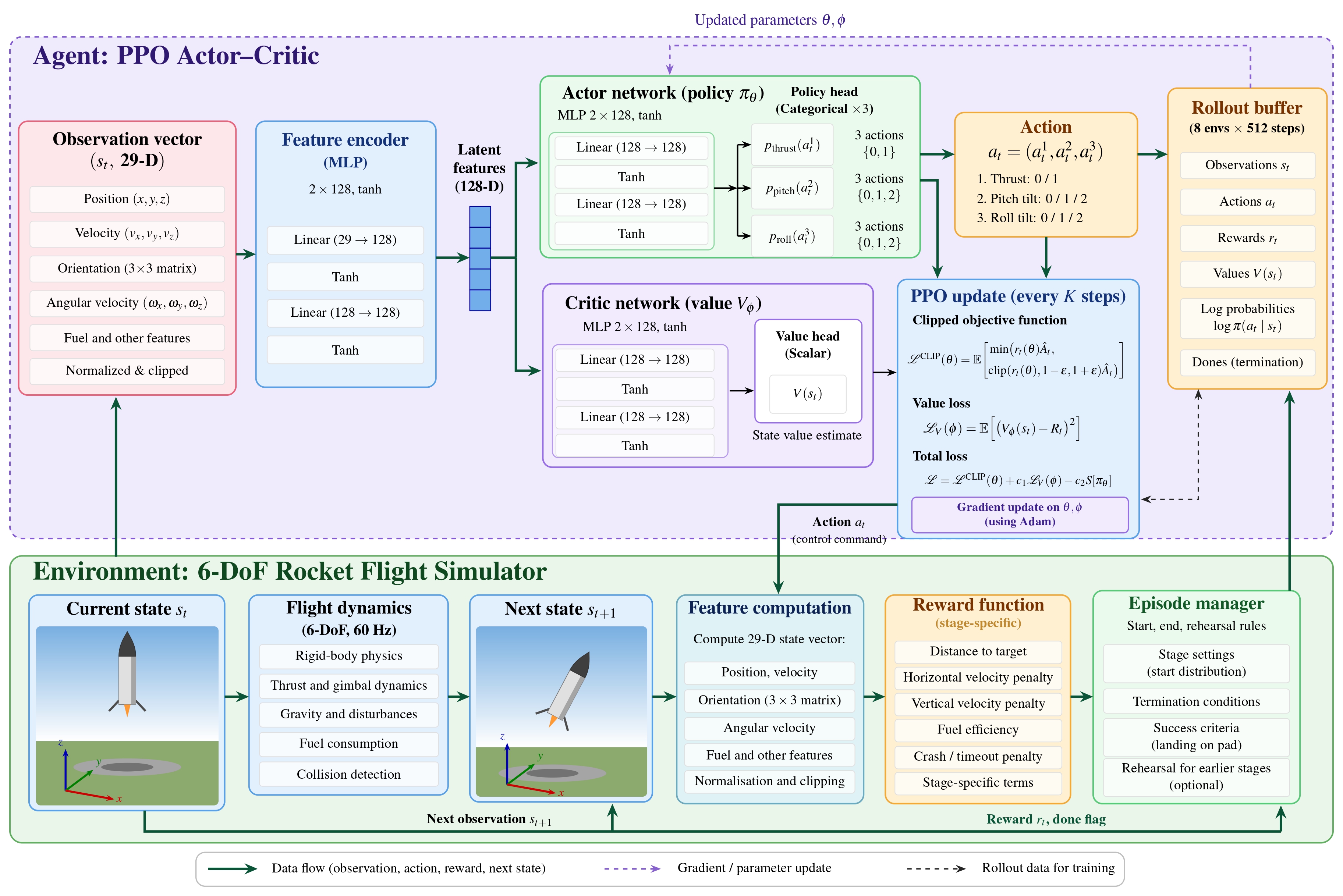}
\caption{Architecture of the curriculum-driven PPO actor-critic agent interacting with the 6-DoF rocket flight simulator. The top block (purple) depicts the neural network topology, action selection, and PPO update mechanism. The bottom block (green) depicts the physics environment, feature computation, and stage-specific reward evaluation. Solid arrows denote data flow (interaction), dashed arrows denote gradient/parameter updates, and dotted arrows denote rollout data.}
\label{fig:architecture}
\end{figure}

The overall system is structured as a closed-loop interaction between the reinforcement learning agent and the physics-based flight simulator. Figure~\ref{fig:architecture} illustrates the complete architecture, detailing the flow of observations, actions, rewards, and gradient updates between the two primary components.

\subsection{The 6-DoF Flight Simulator (Environment)}

The bottom section of the figure depicts the environment, a custom 6-DoF rocket flight simulator. The simulation begins with a current state $s_t$, which includes the rocket's position, velocity, orientation, angular velocity, and fuel levels. The flight dynamics module advances the state by integrating rigid-body physics, thrust and gimbal dynamics, and gravity over a fixed time step. The simulator also models fuel consumption and checks for collisions with the ground or pad. The resulting next state $s_{t+1}$ is then passed through a feature computation module, which constructs the observation vector fed to the agent. This vector is carefully normalized and clipped to ensure numerical stability. The observation is divided into several categories: position, velocity, a $3\times3$ orientation matrix, angular velocity, and other features like fuel and braking margins. Finally, a reward function module evaluates the transition by applying the stage-specific reward terms discussed in Section~4. An episode manager oversees the rollout, determining termination conditions, success criteria, and whether the current episode is a standard stage rollout or a rehearsal of an earlier stage.

\subsection{The PPO Actor-Critic Agent}

The top section of the figure details the PPO agent. The agent receives the observation vector $s_t$ and passes it through a feature encoder, which consists of two linear layers of 128 units with $\tanh$ activations. The encoded latent features are then fed into two separate networks: the actor (policy) and the critic (value function). The actor network also uses two hidden layers of 128 units and outputs a categorical distribution over the 18 joint actions. These actions correspond to the thrust command (0 or 1), pitch tilt (no tilt, tilt forward, tilt backward), and roll tilt (no tilt, tilt left, tilt right). The critic network, sharing the same architecture, outputs a scalar value $V(s_t)$ estimating the expected return from the current state. The selected action $a_t$ is then sent to the simulator, and the transition tuple $(s_t, a_t, r_t, s_{t+1})$ is stored in a rollout buffer. Once a fixed number of steps (or episodes) are collected, the PPO update module computes the clipped surrogate objective, the value loss, and the total loss. Gradients are then computed and used to update the network parameters $\theta$ and $\phi$ via the Adam optimizer. This entire loop repeats, driving the agent to refine its policy and value estimates over time.

\section{Training Methodology}

The success of the reinforcement learning framework relies heavily on the training procedure and the strategic division of the learning task. To address the complexities of far-range navigation and precision landing, we implement a carefully structured training methodology. This section details the Proximal Policy Optimization (PPO) hyperparameters, the curriculum structure, the critic warm-up mechanism, and the rehearsal strategy employed to prevent catastrophic forgetting. 

The reinforcement-learning algorithm is Proximal Policy Optimization as implemented in Stable-Baselines3~\cite{raffin2021sb3,schulman2017ppo}. The curriculum is organised in three stages, and the per-stage hyperparameters are shown in Table~\ref{tab:hp}. Stage one trains for $4\times 10^6$ timesteps with a learning rate of $3\times 10^{-4}$. Stage two trains for $2.5\times 10^6$ timesteps with a learning rate of $1.5\times 10^{-4}$. Stage three trains for $8\times 10^6$ timesteps with a learning rate of $1\times 10^{-4}$. All stages use a discount factor of $0.995$ and a linear learning-rate decay to $10\%$. We discuss the choice of $\gamma = 0.995$ further in Section~\ref{sec:discussion}.

\begin{table}[H]
\centering
\small
\caption{Per-stage PPO hyperparameters.}
\label{tab:hp}
\begin{tabular}{lccc}
\toprule
Hyperparameter & Stage one & Stage two & Stage three \\
\midrule
Learning rate & $3\times 10^{-4}$ & $1.5\times 10^{-4}$ & $1\times 10^{-4}$ \\
Discount factor & $0.995$ & $0.995$ & $0.995$ \\
Entropy coefficient & $0.010$ & $0.005$ & $0.003$ \\
GAE lambda & $0.95$ & $0.95$ & $0.95$ \\
Clipping range & $0.2$ & $0.2$ & $0.2$ \\
Rollout per environment & $512$ & $512$ & $512$ \\
Minibatch size & $512$ & $512$ & $512$ \\
Epochs per update & $10$ & $10$ & $10$ \\
Hidden layers & $128\times 128$ & $128\times 128$ & $128\times 128$ \\
Activation & tanh & tanh & tanh \\
Parallel environments & $8$ & $8$ & $8$ \\
Total timesteps & $4$M & $2.5$M & $8$M \\
\bottomrule
\end{tabular}
\end{table}

When the reward function changes between stage one and stage two, the critic must be re-initialised. To prevent the newly initialised critic from destroying the transferred actor, the actor's parameters are frozen for the first $100{,}000$ timesteps of the new stage and only the critic is trained; after the warm-up, the actor is unfrozen. No critic re-initialisation is needed between stage two and stage three, since both stages share the landing reward. To prevent catastrophic forgetting, a fraction of each stage's episodes is drawn from the earlier stages. Stage two mixes in $15\%$ stage-one episodes, and stage three mixes in $10\%$ stage-one and $10\%$ stage-two episodes.

\section{Training Diagnostics}
\label{sec:diagnostics}

This section analyses the training dynamics of the three curriculum stages through reward, episode-length, and critic-conditioning curves recorded during training. The curves shown are the raw scalars exported from TensorBoard; the slight high-frequency noise in the raw traces has been smoothed for readability only in the smoothed overlay, while the underlying raw values are preserved. 
By examining these metrics, we can verify the necessity of the curriculum structure and the critic warm-up procedure, and observe how the agent progressively masters the complex maneuvers required for a successful landing.

\subsection{Stage One: Far Approach}

In the first stage, the agent must learn to navigate a vast distance to reach the landing zone. The training curves for this stage reflect the agent's exploration of the large workspace and its gradual discovery of an efficient trajectory to the hover point.

\begin{figure}[H]
\centering
\begin{subfigure}{0.48\textwidth}
\includegraphics[width=\linewidth]{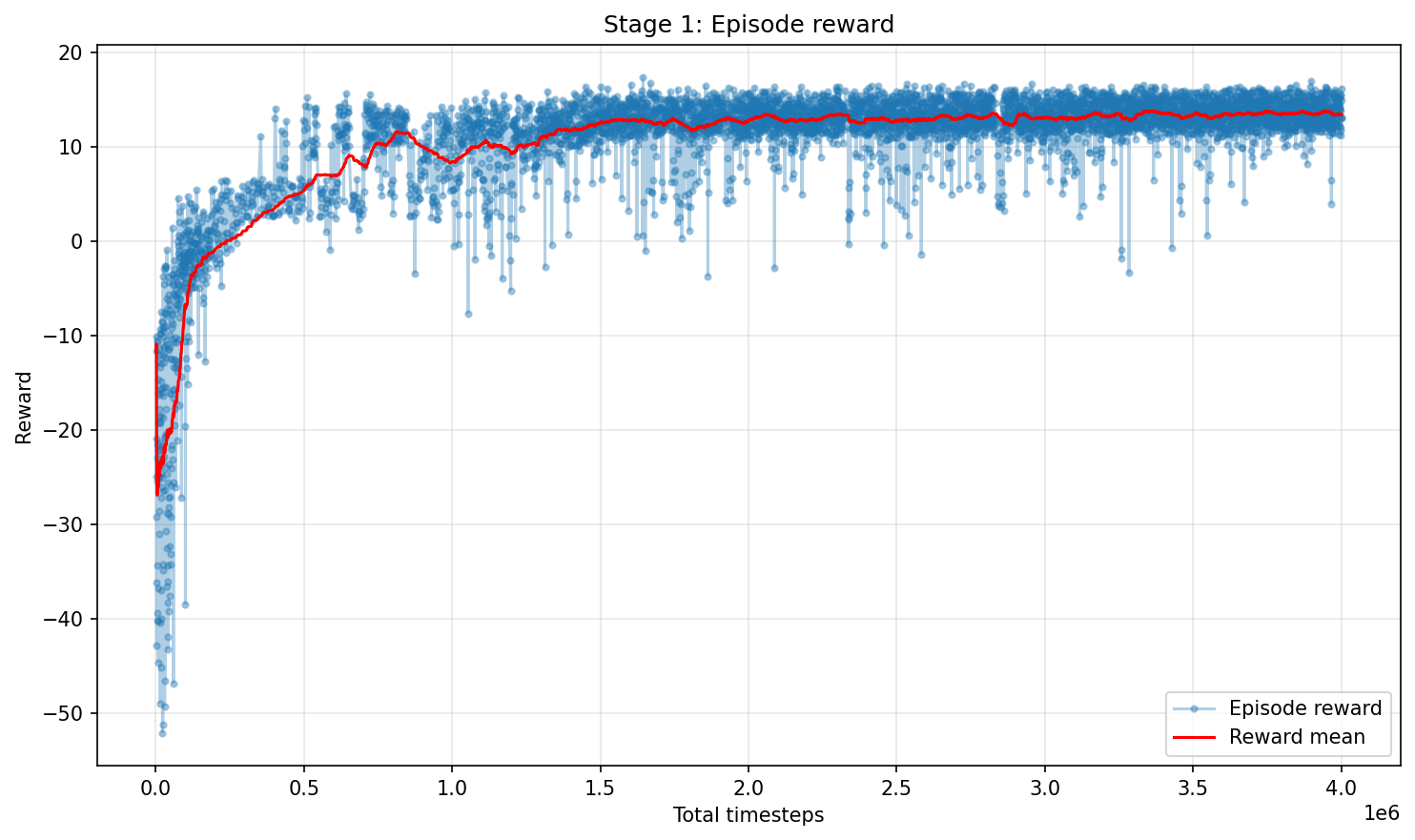}
\caption{Mean episode reward}
\end{subfigure}
\hfill
\begin{subfigure}{0.48\textwidth}
\includegraphics[width=\linewidth]{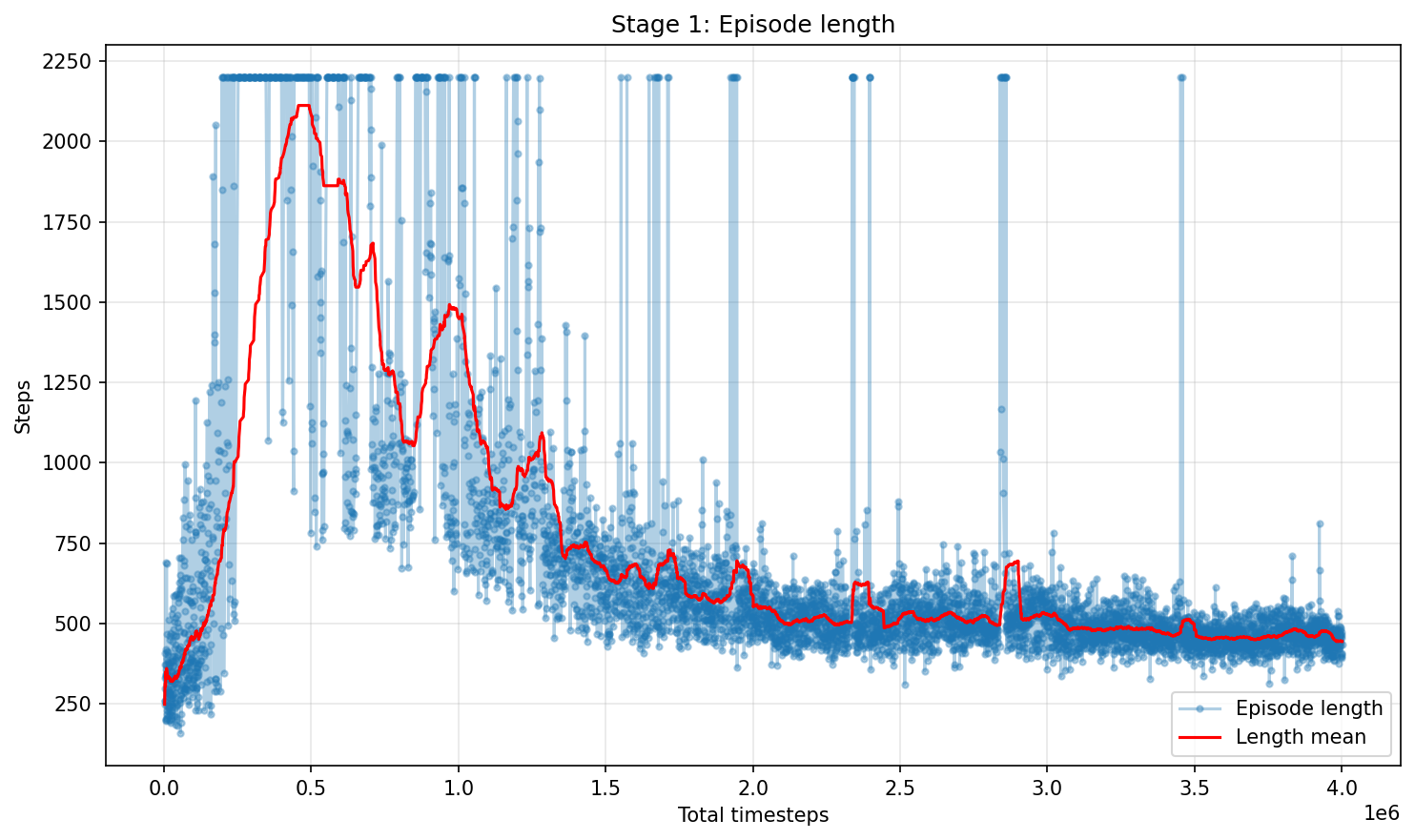}
\caption{Mean episode length (frames)}
\end{subfigure}
\caption{Stage one (far approach) training curves. Left: episode reward rises steadily as the agent learns to navigate the $1{,}000$--$1{,}500$ unit gap to the pad. Right: episode length in physics frames initially spikes as the agent explores the large workspace, then settles as it discovers an efficient trajectory to the hover zone.}
\label{fig:s1_behaviour}
\end{figure}

\begin{figure}[H]
\centering
\begin{subfigure}{0.48\textwidth}
\includegraphics[width=\linewidth]{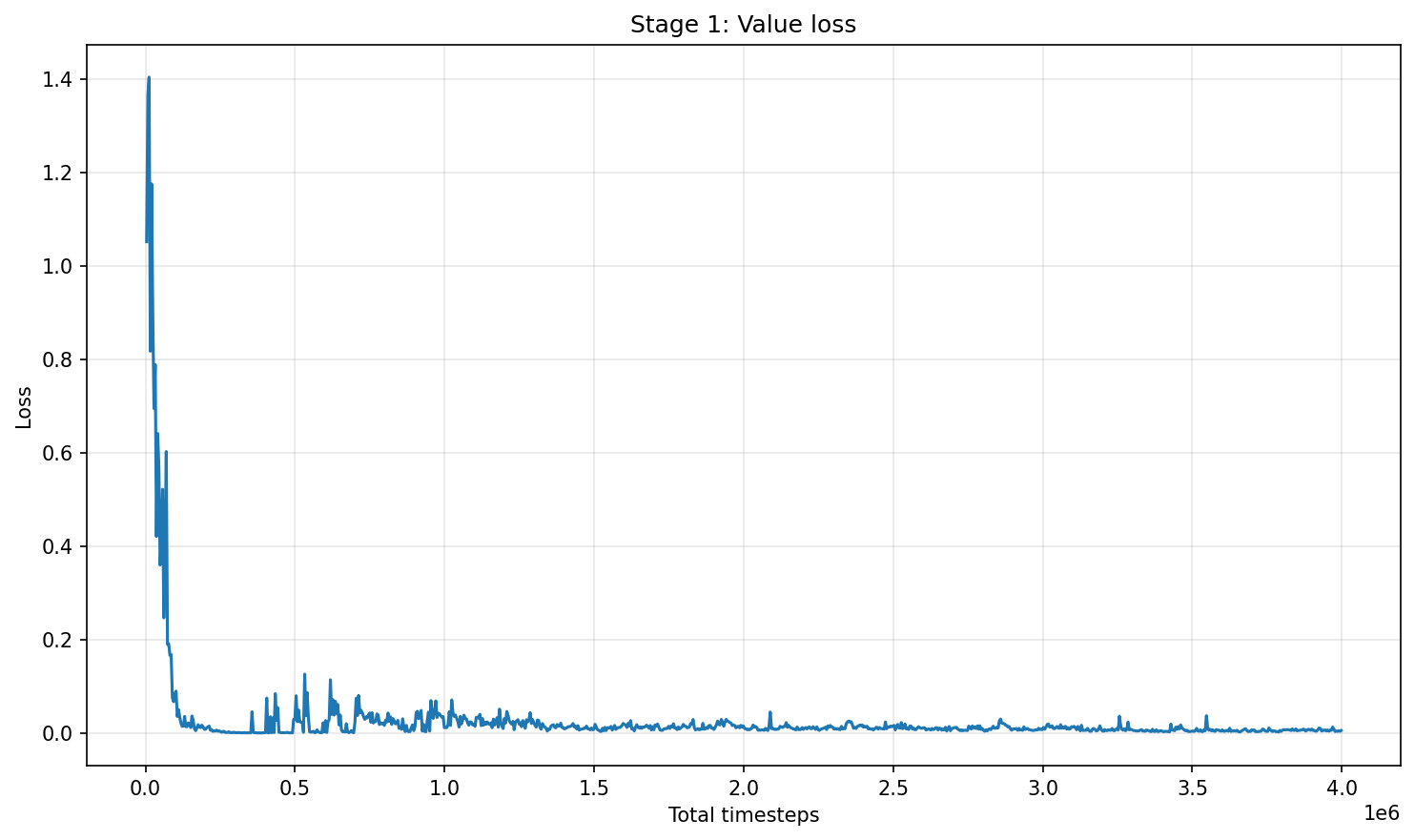}
\caption{Value loss}
\end{subfigure}
\hfill
\begin{subfigure}{0.48\textwidth}
\includegraphics[width=\linewidth]{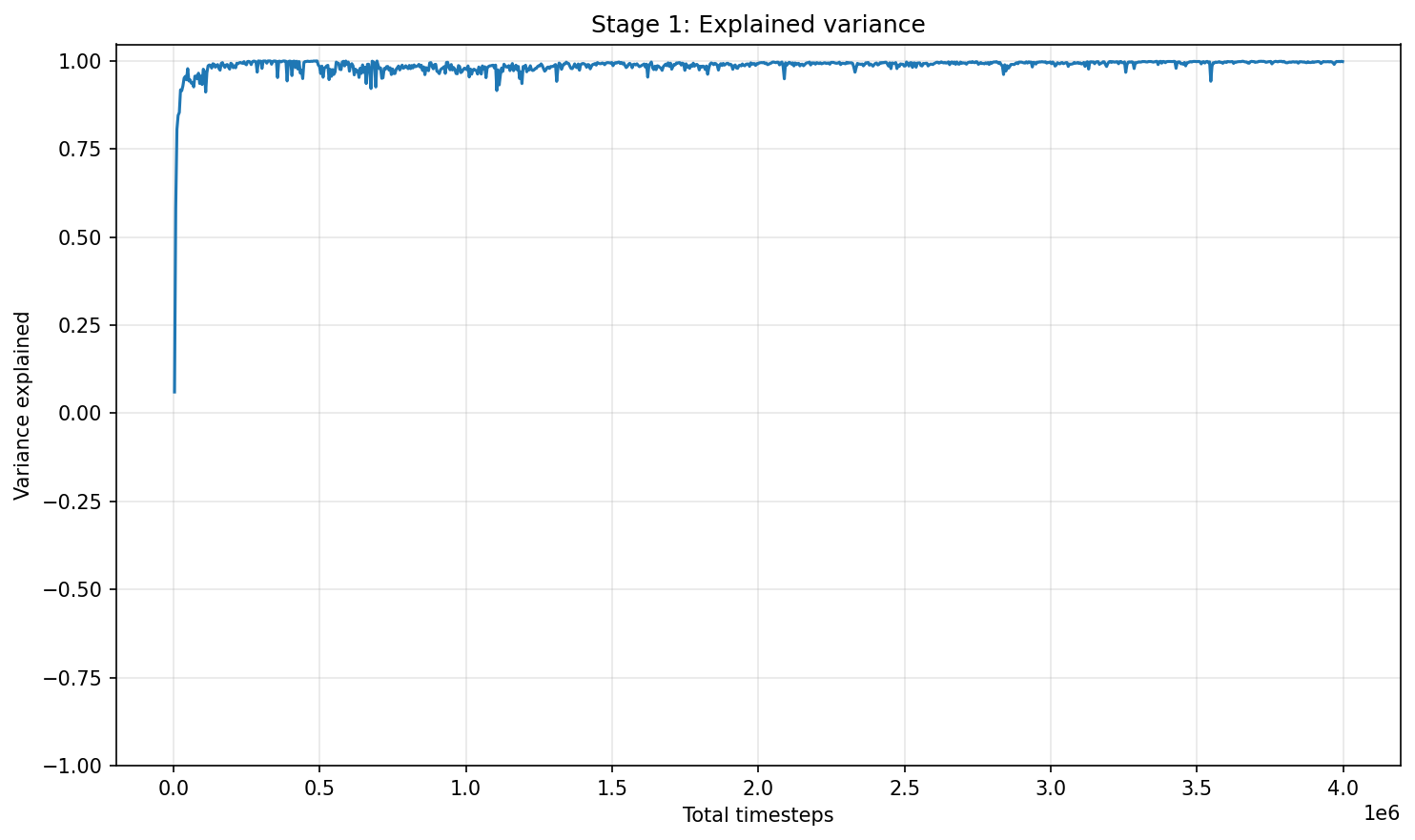}
\caption{Explained variance}
\end{subfigure}
\caption{Stage one critic conditioning. Left: value loss drops sharply within the first million timesteps. Right: explained variance rapidly reaches $1.0$, showing the critic achieves an essentially perfect fit of the value function. This guarantees the actor receives clean, well-conditioned gradient updates at the transfer point.}
\label{fig:s1_critic}
\end{figure}

The stage-one curves in Figures~\ref{fig:s1_behaviour} and~\ref{fig:s1_critic} show the idealised behaviour that the curriculum relies on. The agent learns the far-approach task efficiently, and the critic's estimate of the return becomes exact before the stage ends, so the transferred actor will not be degraded by critic noise during the warm-up window of stage two. The decay in episode length is driven by the hover-zone success criterion described in Section~4.1; the agent learns to satisfy the hover condition sooner as training progresses.

\subsection{Stage Two: Soft Descent}

The second stage focuses on the vertical descent and braking. The agent starts near the pad and must learn to slow its descent to a safe speed for touchdown. The training curves here highlight the impact of the reward function change and the effectiveness of the critic warm-up.

\begin{figure}[H]
\centering
\begin{subfigure}{0.48\textwidth}
\includegraphics[width=\linewidth]{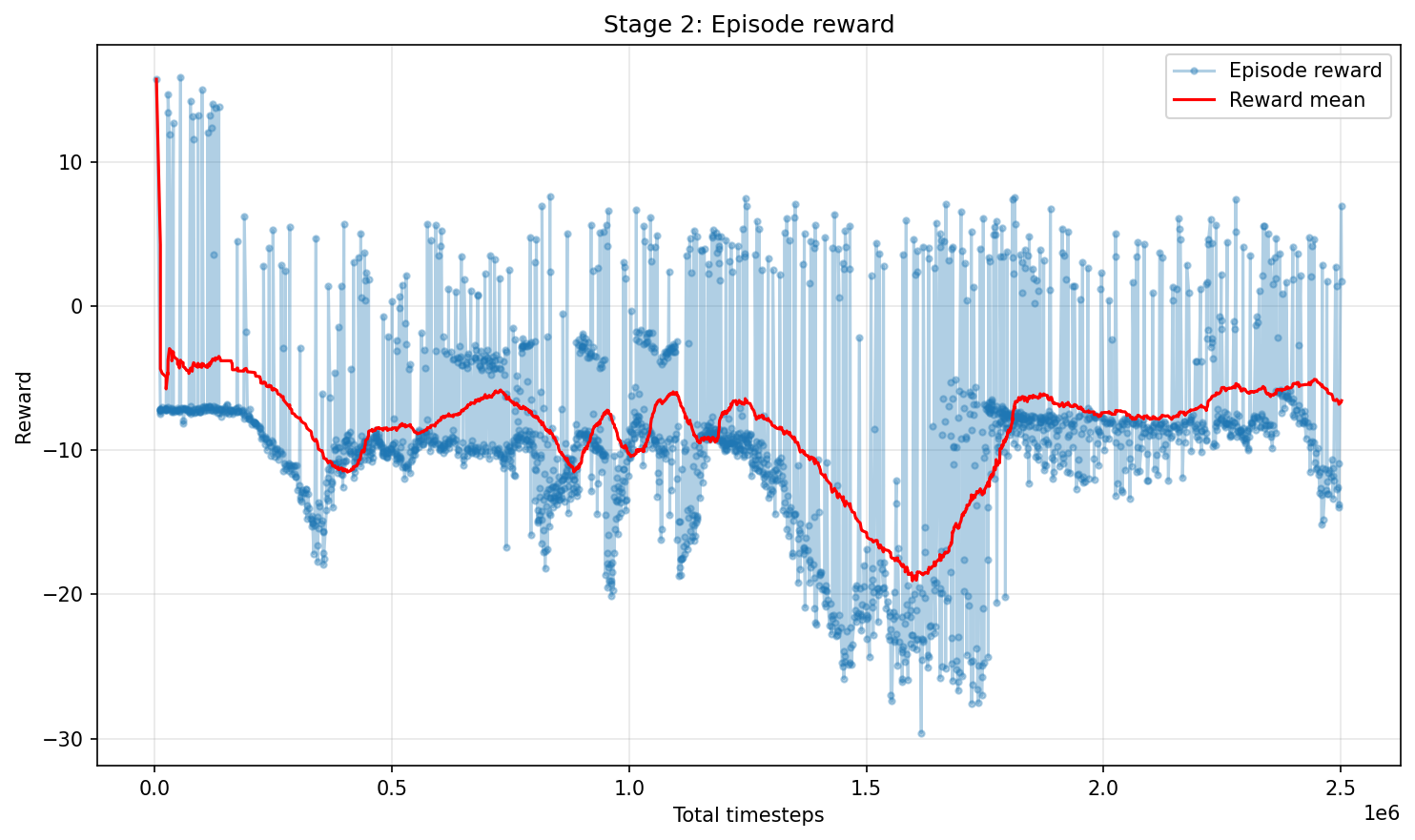}
\caption{Mean episode reward}
\end{subfigure}
\hfill
\begin{subfigure}{0.48\textwidth}
\includegraphics[width=\linewidth]{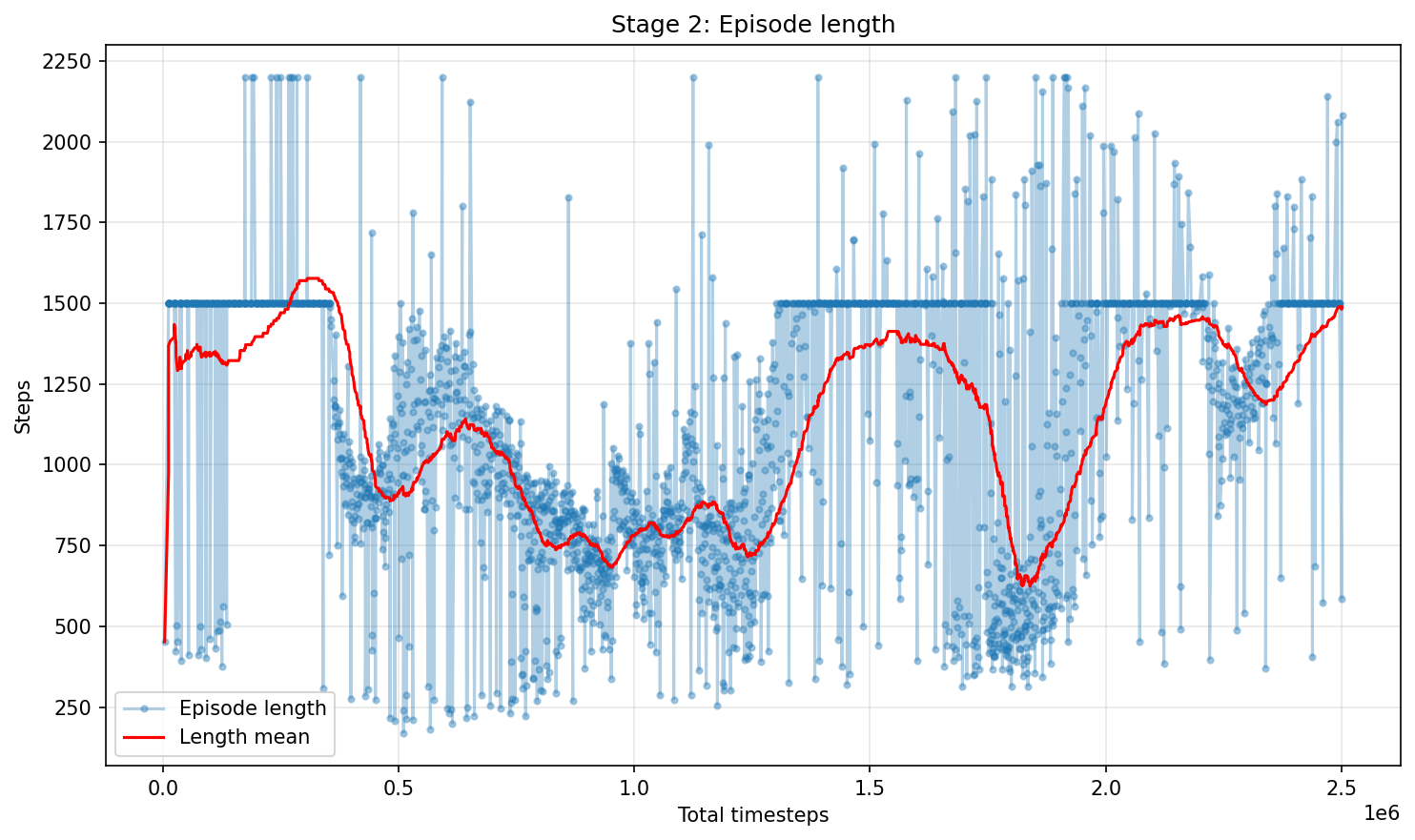}
\caption{Mean episode length (frames)}
\end{subfigure}
\caption{Stage two (soft descent) training curves. Left: reward initially dips as the reward function changes, then recovers and rises after the $100{,}000$-timestep critic warm-up. Right: episode length stabilises as the agent learns the vertical braking profile.}
\label{fig:s2_behaviour}
\end{figure}

\begin{figure}[H]
\centering
\begin{subfigure}{0.48\textwidth}
\includegraphics[width=\linewidth]{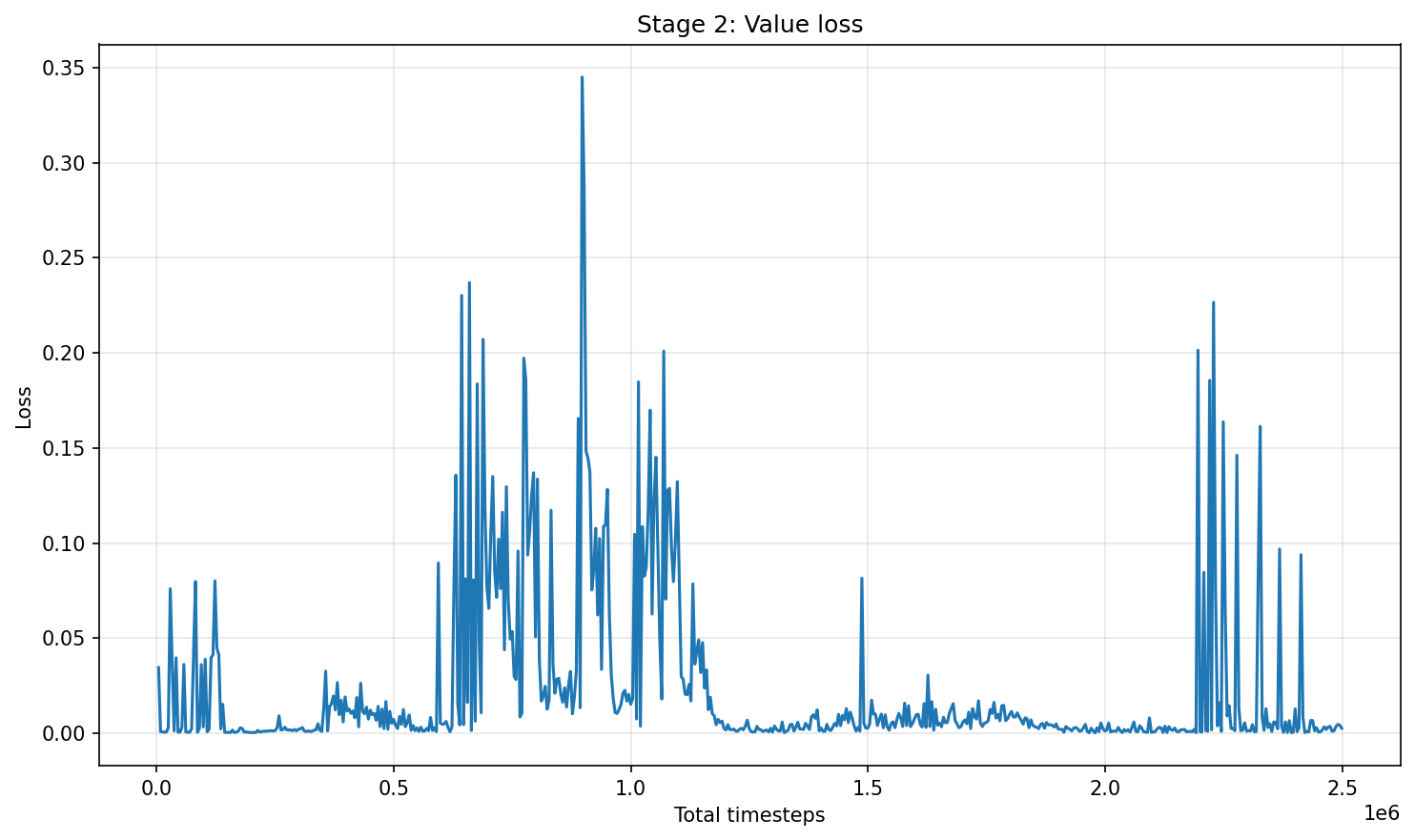}
\caption{Value loss}
\end{subfigure}
\hfill
\begin{subfigure}{0.48\textwidth}
\includegraphics[width=\linewidth]{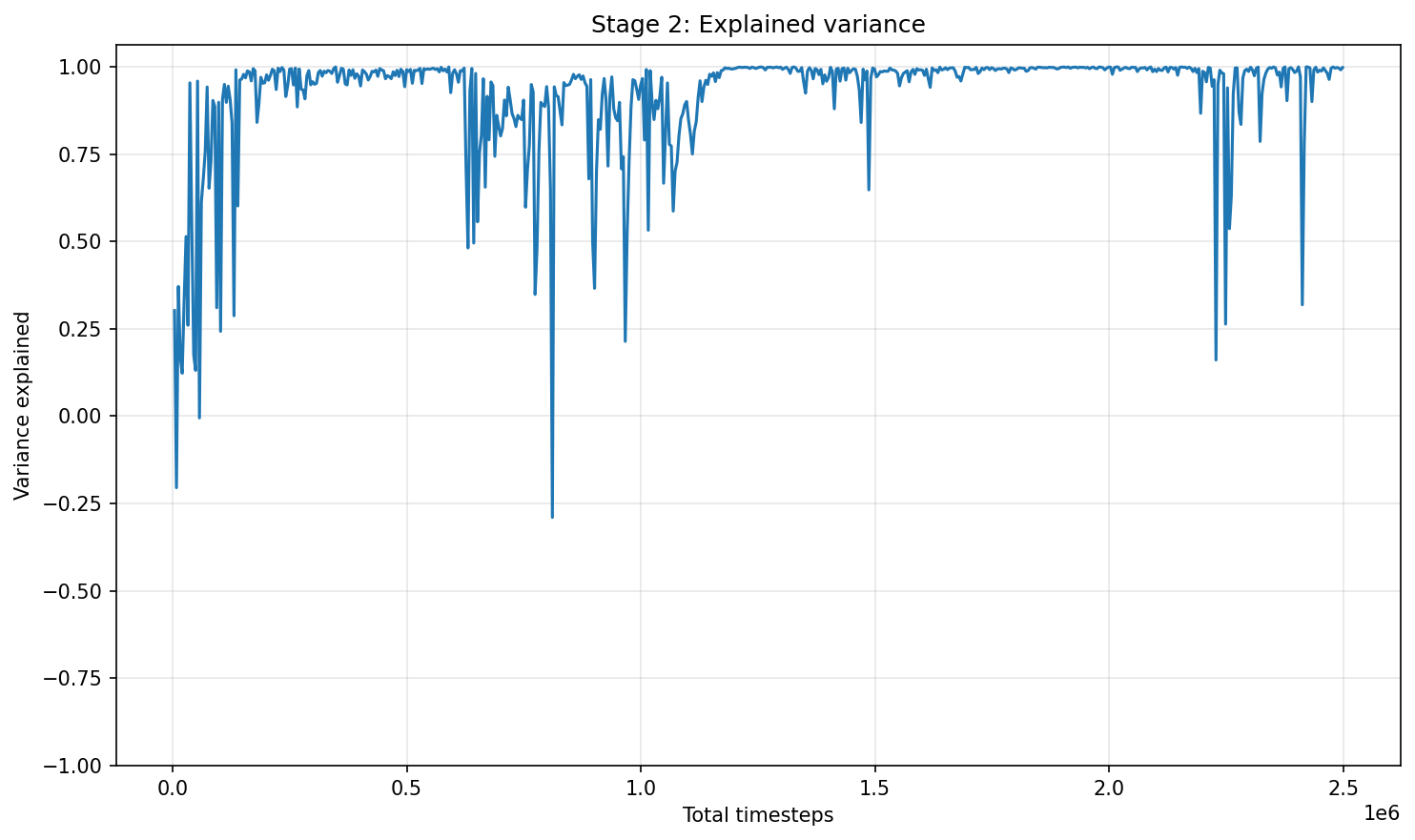}
\caption{Explained variance}
\end{subfigure}
\caption{Stage two critic conditioning. Left: the value loss exhibits initial spikes because the critic is freshly initialised for the new reward function. This justifies the critic warm-up. Right: explained variance recovers quickly as the critic adapts to the landing objective.}
\label{fig:s2_critic}
\end{figure}

The stage-two curves illustrate the necessity of the critic warm-up. When the reward changes at the start of stage two, the value loss spikes dramatically because the critic no longer understands the environment. The rehearsal mechanism also introduces variance, as the agent occasionally sees stage-one episodes. As the warm-up proceeds and the critic recovers, the reward curve begins its steady climb.

\subsection{Stage Three: Full Task}

In the final stage, the agent must integrate both the far-range navigation and the soft descent into a single, coherent policy. The training curves show the agent's struggle and eventual success in balancing these competing objectives.

\begin{figure}[H]
\centering
\begin{subfigure}{0.48\textwidth}
\includegraphics[width=\linewidth]{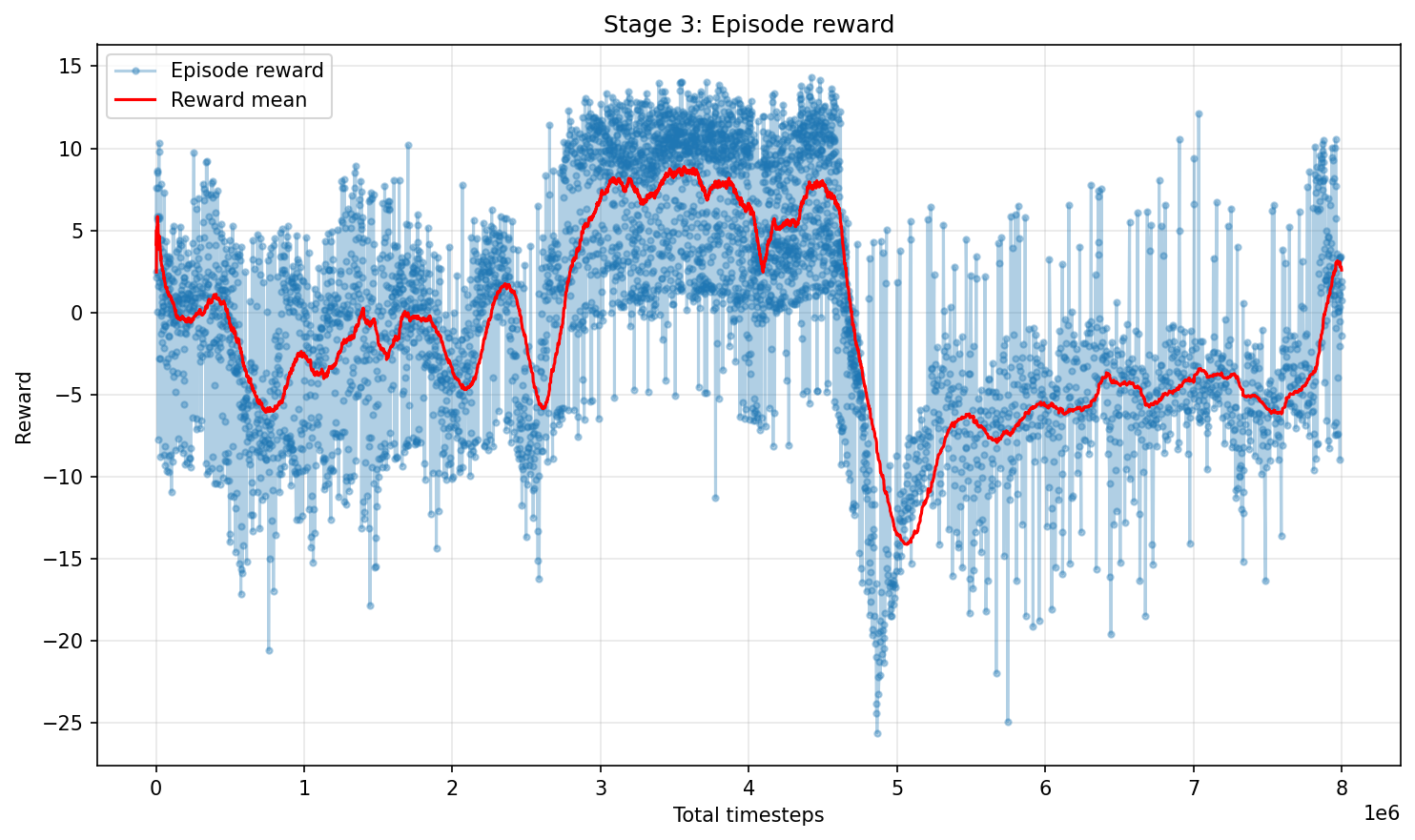}
\caption{Mean episode reward}
\end{subfigure}
\hfill
\begin{subfigure}{0.48\textwidth}
\includegraphics[width=\linewidth]{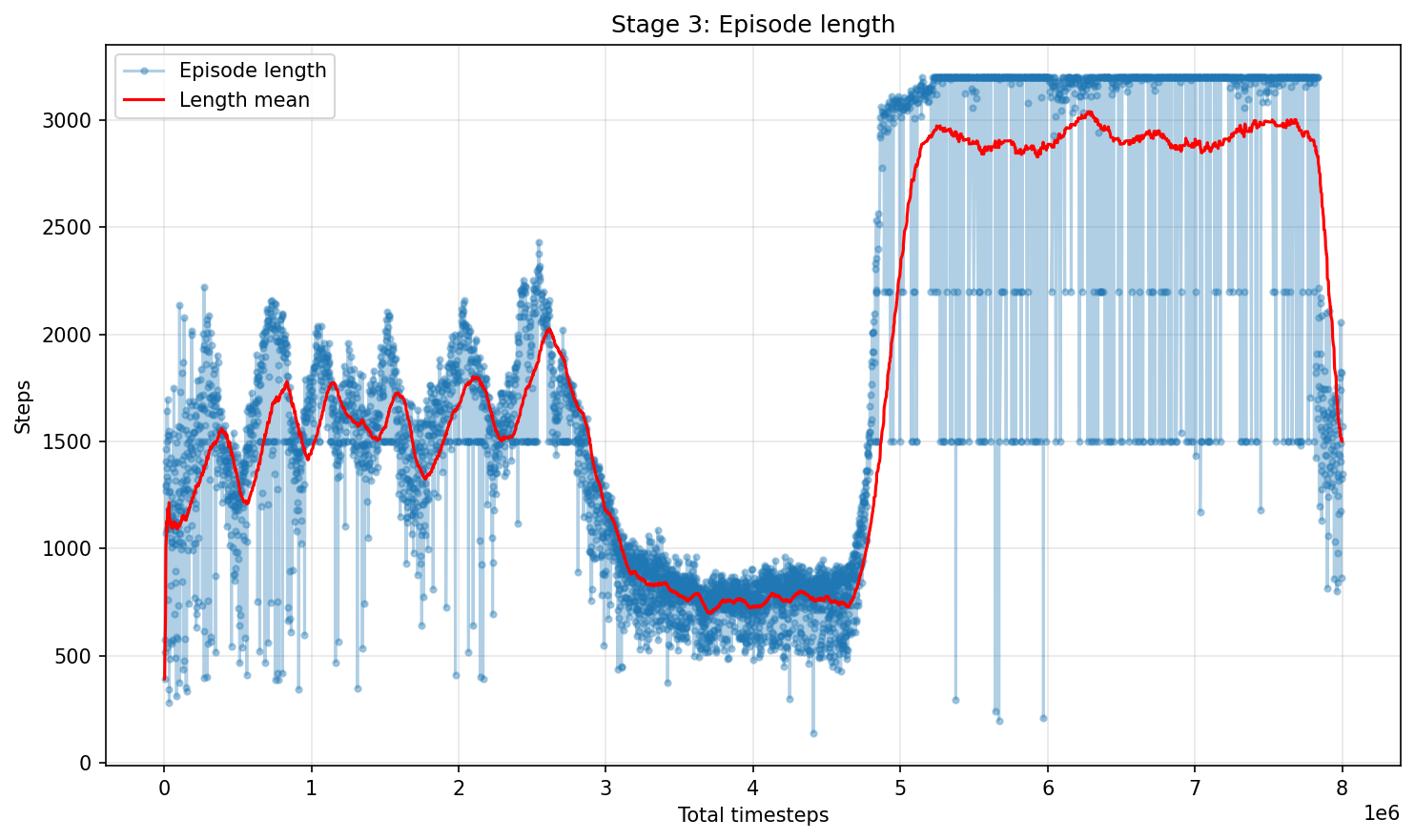}
\caption{Mean episode length (frames)}
\end{subfigure}
\caption{Stage three (full task) training curves. Left: reward stabilises as the agent learns the composite navigation and landing task. Right: episode length becomes highly variable as the agent alternates between the long approach and the soft descent, reflecting the rehearsal mechanism.}
\label{fig:s3_behaviour}
\end{figure}

\begin{figure}[H]
\centering
\begin{subfigure}{0.48\textwidth}
\includegraphics[width=\linewidth]{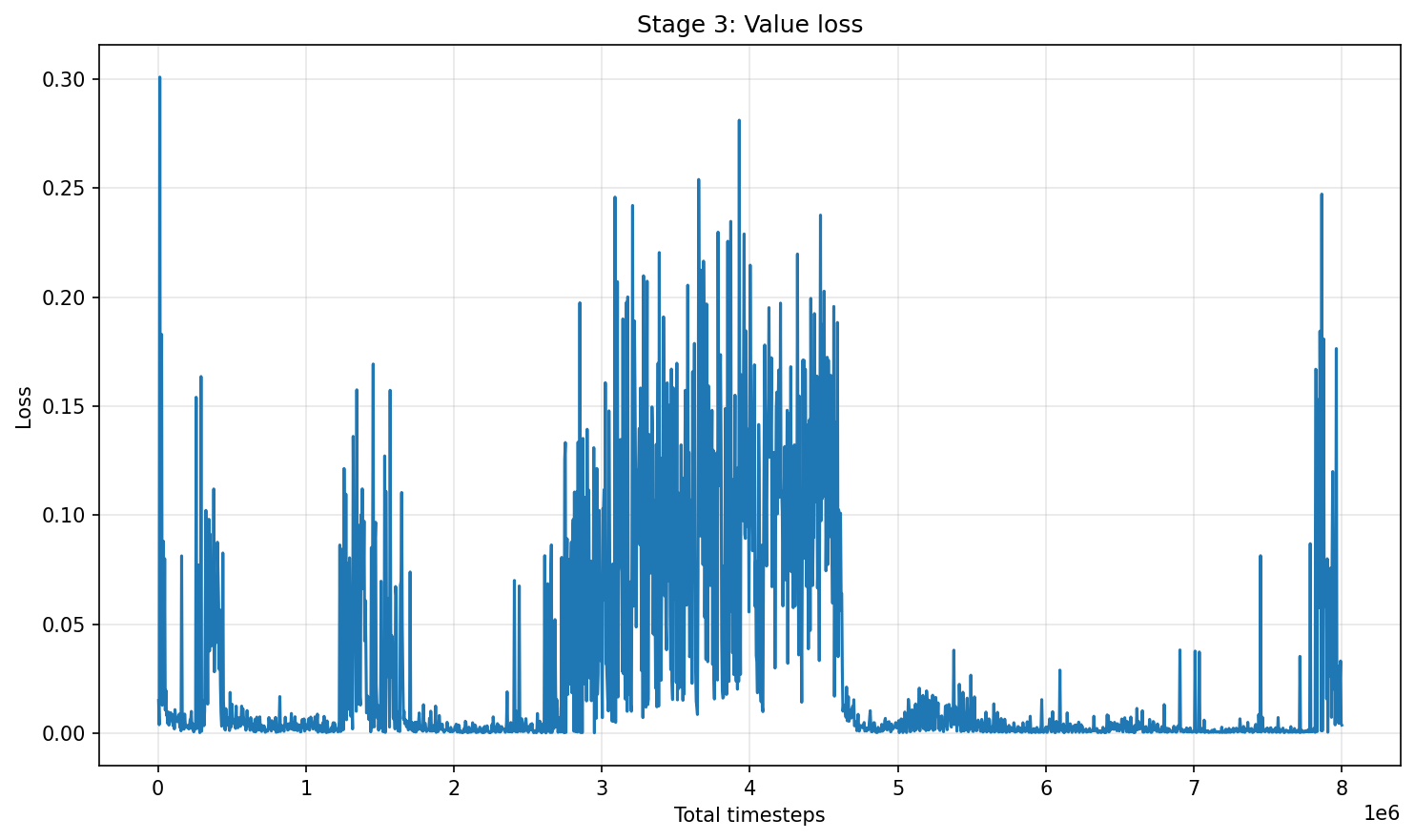}
\caption{Value loss}
\end{subfigure}
\hfill
\begin{subfigure}{0.48\textwidth}
\includegraphics[width=\linewidth]{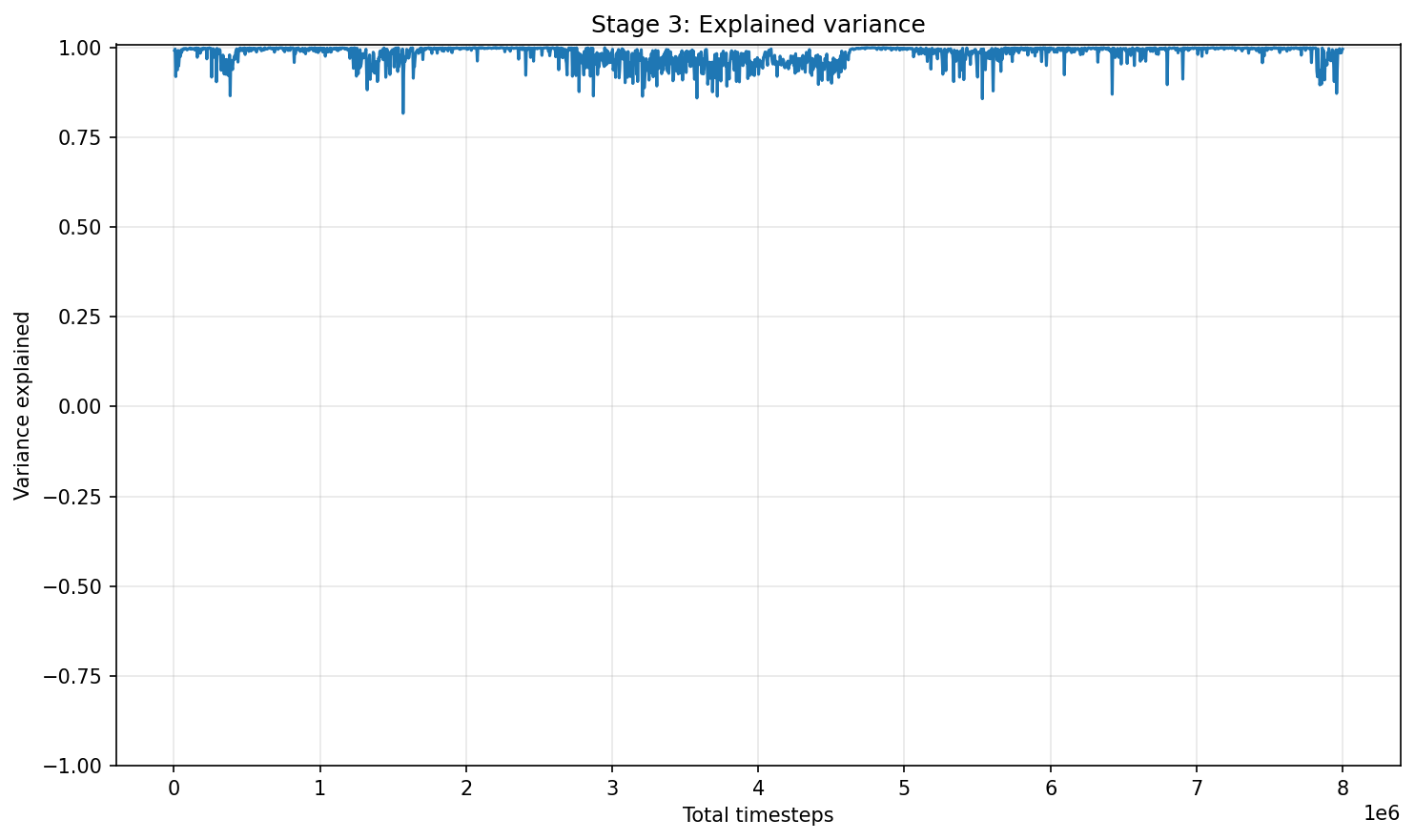}
\caption{Explained variance}
\end{subfigure}
\caption{Stage three critic conditioning. No warm-up is required because stage three inherits the same reward and a well-conditioned critic from stage two. The value loss remains extremely low and the explained variance remains near $1.0$, allowing the policy to focus on the composite manoeuvre.}
\label{fig:s3_critic}
\end{figure}

The stage-three curves in Figures~\ref{fig:s3_behaviour} and~\ref{fig:s3_critic} show the full-task dynamics. Because the reward function and the critic transfer directly from stage two, the critic conditioning is excellent from timestep zero. The episode length varies due to the rehearsal mechanism, but the overall reward stabilises as the policy successfully integrates the far approach with the soft descent.

\section{Experimental Setup and Results}
\label{sec:results}

This section outlines the experimental setup and presents the evaluation results. It describes the initial conditions and implemented success definition, then compares the landing outcomes with the baseline study.

\subsection{Evaluation Protocol}

The training budget is $4\times 10^6$ timesteps for stage one, $2.5\times 10^6$ for stage two, and $8\times 10^6$ for stage three, giving a total of $14.5\times 10^6$ timesteps. The stage-three environment samples the initial horizontal distance from the pad centre uniformly from $1{,}000$ to $1{,}500$ units. For the coordinate-wise distance summaries in Table~\ref{tab:stats}, we report the absolute initial offsets $|x_0|$ and $|y_0|$, obtained by taking the absolute value of each episode's initial $x$ and $y$ coordinates before calculating the sample mean and standard deviation. These two quantities describe distances along the coordinate axes and are nonnegative. They are distinct from the radial horizontal distance $d_0=\sqrt{x_0^2+y_0^2}$, which measures the rocket's straight-line horizontal distance from the pad centre and is reported separately.


The evaluation summary contains episode-level measurements of the initial and final states, elapsed physics frames, and landing quantities. Table~\ref{tab:stats} summarises the reported per-field statistics over the \(100{,}000\)-episode evaluation, in the form of sample mean and sample standard deviation. The four-condition result is calculated from the touchdown measurements; this explicitly checks the $10^\circ$ tilt threshold even though the saved Stage~3 configuration has \texttt{require\_upright\_touchdown=False}. Thus, the tilt result represents observed touchdown compliance, rather than a tilt constraint enforced by the Stage~3 success flag.

\begin{table}[H]
\centering
\small
\caption{Evaluation statistics over \(100{,}000\) episodes. Values are sample mean \(\pm\) sample standard deviation. The initial $x$- and $y$-distance rows report absolute coordinate offsets, $|x_0|$ and $|y_0|$; the initial horizontal distance row reports radial distance from the pad centre. Lengths are in simulator units and frames.}
\label{tab:stats}
\begin{tabular}{lcc}
\toprule
Field & Mean \(\pm\) std & Count \\
\midrule
Success & \(0.99980 \pm 0.01414\) & \(100{,}000\) \\
Frames & \(2955.91 \pm 261.02\) & \(100{,}000\) \\
Within pad radius & \(0.99980 \pm 0.01414\) & \(100{,}000\) \\
Vertical-speed condition satisfied & \(1.00000 \pm 0.00000\) & \(100{,}000\) \\
Horizontal-speed condition satisfied & \(1.00000 \pm 0.00000\) & \(100{,}000\) \\
\midrule
Initial $x$-distance, $|x_0|$ & \(795.996 \pm 397.713\) & \(100{,}000\) \\
Initial $y$-distance, $|y_0|$ & \(796.415 \pm 397.129\) & \(100{,}000\) \\
Initial horizontal distance from pad centre, $d_0$ & \(1250.188 \pm 144.222\) & \(100{,}000\) \\
Initial \(z\) & \(1099.698 \pm 115.166\) & \(100{,}000\) \\
Initial \(v_z\) & \(-0.84947 \pm 0.37487\) & \(100{,}000\) \\
Initial tilt & \(15.2706^\circ \pm 5.6410^\circ\) & \(100{,}000\) \\
\midrule
Final \(x\) & \(3.7494 \pm 1.9911\) & \(100{,}000\) \\
Final \(y\) & \(5.6234 \pm 2.0635\) & \(100{,}000\) \\
Final \(z\) & \(12.9472 \pm 0.0316\) & \(100{,}000\) \\
Final \(v_z\) & \(-0.10567 \pm 0.01405\) & \(100{,}000\) \\
Final tilt & \(0.3955^\circ \pm 0.1957^\circ\) & \(100{,}000\) \\
Distance from pad centre & \(7.0673 \pm 1.9893\) & \(100{,}000\) \\
Horizontal speed & \(0.01256 \pm 0.00675\) & \(100{,}000\) \\
Fuel left & \(691.151 \pm 26.827\) & \(100{,}000\) \\
\bottomrule
\end{tabular}
\end{table}

\subsection{Results}

The comparison comprises $100{,}000$ evaluation episodes. The proposed policy and the deterministic policy from the baseline of Chowdhury et al.~\cite{chowdhury2024icaart} were evaluated on the test set. On that test set, the proposed policy achieved $99{,}980$ successful episodes, corresponding to a success rate of $99.98\%$, whereas the baseline achieved $68.31\%$. These success-rate figures refer to the test set described in the Evaluation Protocol.

It is crucial to emphasise the disparity in the initial conditions between the two methods. In the baseline study, the rocket starts at a very low distance from the landing pad, positioned almost directly above it, meaning the agent is not required to perform any significant lateral navigation. Despite its near-pad starting conditions, the baseline achieved successful landings in $68.31\%$ of episodes in the comparison. In stark contrast, the proposed method starts the rocket between $1{,}000$ and $1{,}500$ units away from the landing pad, forcing the agent to execute a complex, multi-phase lateral approach before it can even begin the vertical descent. Despite this vastly harder initial condition, the proposed method achieves a near-perfect accuracy of $99.98\%$. This strongly indicates that the curriculum, the observation redesign, and the reward redesign are fundamentally more capable of capturing the underlying physics than the flat training approach of the baseline.

Table~\ref{tab:compare} gives the qualitative head-to-head comparison of the design choices.

\begin{table}[H]
\centering
\small
\caption{Design-level comparison with the study of Chowdhury et al.~\cite{chowdhury2024icaart}.}
\label{tab:compare}
\begin{tabular}{lcc}
\toprule
Metric & Baseline of~\cite{chowdhury2024icaart} & Proposed work \\
\midrule
Training paradigm & flat single-stage & three-stage curriculum with rehearsal \\
Attitude representation & quaternion plus angular-rate proxies & full $3\times 3$ rotation matrix \\
Observation dimension & $13$ & $29$ \\
Action space & $6$ & $18$, joint $2\times 3\times 3$ \\
Reward terms & $8$ shaping plus $1$ compound terminal & $8$ shaping plus $3$ explicit terminal \\
Reward constants & $15$ & $8$ \\
Reward clipping & interval $[-500,\,+500]$ & per-step return scaled by $0.1$ \\
Actor and critic hidden layers & $256\times 256\times 256$ & $128\times 128$ \\
Parameter count, actor and critic & $\approx 273{,}000$ & $\approx 40{,}000$ \\
Discount factor & $0.85$ & $0.995$ \\
Entropy schedule & linear anneal & per-stage tuned \\
Critic warm-up on transfer & no & $100{,}000$ timesteps \\
Rehearsal mechanism & no & $10$--$15\%$ earlier stage episodes \\
Transfer between stages & no & stage $1\to 2\to 3$ \\
Total timesteps & $10$M & $14.5$M \\
Initial horizontal distance & very small (near pad) & $1{,}000$ to $1{,}500$ units \\
\bottomrule
\end{tabular}
\end{table}

Table~\ref{tab:100k_compare} presents the quantitative comparison for the $100{,}000$-episode test set. The baseline achieved a $68.31\%$ success rate on that test set, compared with $99.98\%$ for the proposed policy. The reported touchdown measurements also show differences in mean tilt, vertical speed, and horizontal speed between the two policies. 

In the far-range evaluation, the proposed method leaves \(691\) units out of $1{,}000$ (approximately \(69.1\%\)) of its initial fuel capacity.

\begin{table}[H]
\centering
\small
\caption{Quantitative comparison of reported landing outcomes between the baseline of Chowdhury et al.~\cite{chowdhury2024icaart} and the proposed method over the \(100{,}000\)-episode test set.}
\label{tab:100k_compare}
\begin{tabular}{lccc}
\toprule
Metric & Baseline of~\cite{chowdhury2024icaart} & Proposed work & Improvement \\
\midrule
Success rate & \(68.31\%\) & \(99.98\%\) & \(+31.67\) pp \\
Touchdown tilt angle $\theta$ & \(11.2^\circ\) & \(0.396^\circ \pm 0.196^\circ\) & \(\approx 28.3\times\) lower \\
Touchdown vertical speed \(|v_z|\) & \(0.190 \pm 0.045\) & \(0.106 \pm 0.014\) & \(\approx 1.8\times\) lower \\
Distance from pad centre & \(13.10 \pm 5.1\) & \(7.07 \pm 1.99\) & \(\approx 1.9\times\) lower \\
Touchdown horizontal speed & \(0.152 \pm 0.021\) & \(0.0126 \pm 0.0068\) & \(\approx 12\times\) lower \\
\bottomrule
\end{tabular}
\end{table}

The proposed framework's \(99.98\%\) success over \(100{,}000\) episodes yields a \(95\%\) Wilson confidence interval of \(\approx [99.97\%,\,99.99\%]\). The baseline's \(68.31\%\) over the same number of episodes yields an interval of \(\approx [68.0\%,\,68.6\%]\). The two intervals do not overlap, so the observed improvement is statistically decisive at the $95\%$ level of the evaluation-sampling error. The baseline's original paper reported \(84\%\) success over \(100\) episodes; the comparison here reports \(68.31\%\) on the test set described above.

The horizontal performance is the dimension in which the task most differs from the baseline. The initial radial horizontal distance from the pad centre is sampled uniformly from $1{,}000$ to $1{,}500$ units, which is $\approx 78\times$ the pad radius. The mean final distance from pad centre of \(7.07 \pm 1.99\) units reported in Table~\ref{tab:post} therefore represents a mean horizontal correction of roughly \(1243\) units per episode. The baseline achieves a notably worse final distance (\(13.10 \pm 5.1\)), despite beginning almost directly above the pad. Its inability to traverse the long distance from the wider initial distribution, and its imprecision even when close, are precisely what cause its success rate to collapse to \(68.31\%\).

\begin{table}[H]
\centering
\small
\caption{Average measured landing parameters for the proposed method across the evaluation episodes. For quantities with a stated operating limit, the last column gives the fraction of that limit represented by the reported mean.}
\label{tab:post}
\begin{tabular}{lccc}
\toprule
Quantity & Mean $\pm$ standard deviation & Limit & Fraction of budget used \\
\midrule
Touchdown tilt angle $\theta$ & \(0.396^\circ \pm 0.196^\circ\) & \(\le 10^\circ\) & \(\approx 3.95\%\) \\
Touchdown vertical speed \(|v_z|\) & \(0.106 \pm 0.014\) & \(\le 0.15\) & \(\approx 70.4\%\) \\
Distance from pad centre & \(7.07 \pm 1.99\) & \(\le 16\) & \(\approx 44.2\%\) \\
Touchdown horizontal speed & \(0.0126 \pm 0.0068\) & \(\le 0.25\) & \(\approx 5.0\%\) \\
Fuel remaining & \(691.15 \pm 26.83\) & \(1{,}000\) max & \(\approx 69.1\%\) of capacity \\
\bottomrule
\end{tabular}
\end{table}

Table~\ref{tab:four_conditions} reports the proportion of episodes satisfying each touchdown condition individually and all four conditions simultaneously. Every touchdown satisfies the tilt, vertical-speed, and horizontal-speed thresholds; the 20 episodes that do not satisfy the joint criterion are those whose distance from the pad centre exceeds the pad radius.

\begin{table}[H]
\centering
\small
\caption{Four-condition touchdown assessment for the $100{,}000$-episode evaluation. Each condition is checked from the touchdown state.}
\label{tab:four_conditions}
\begin{tabular}{lcc}
\toprule
Touchdown condition & Episodes satisfying condition & Rate \\
\midrule
Tilt angle $\theta \leq 10^\circ$ & $100{,}000$ & $100.00\%$ \\
Distance from pad centre $d_h \leq 16$ & $99{,}980$ & $99.98\%$ \\
Vertical velocity $v_z \geq -0.15$ units/frame & $100{,}000$ & $100.00\%$ \\
Horizontal speed $v_h \leq 0.25$ units/frame & $100{,}000$ & $100.00\%$ \\
\midrule
All four conditions simultaneously & $99{,}980$ & $99.98\%$ \\
\bottomrule
\end{tabular}
\end{table}

```latex
\section{Discussion}
\label{sec:discussion}

The training curves in Section~\ref{sec:diagnostics} clarify why the curriculum succeeds where flat training would fail. In stage one the reward landscape is dominated by horizontal progress and alignment, and PPO converges in $4\times 10^6$ timesteps to a policy that navigates the $1{,}000$--$1{,}500$ unit gap to the hover point. In stage two the actor and critic are transferred from stage one, and the critic warm-up of \(100{,}000\) timesteps prevents the newly initialised critic from corrupting the transferred actor's policy gradient. In stage three the transfer is direct, and the rehearsal mechanism ensures the agent does not forget the far-approach or soft-descent skills while learning the composite task.

The observation change is responsible for the small network size. The high-dimensional vector explicitly provides the horizontal distance, radial velocity, and braking margins, allowing the agent to plan its approach. Because the observation exposes the quantities that the action channels actually modulate, a two-layer \(128\)-unit network suffices where the baseline needed three layers of \(256\) units.

The reward redesign is responsible for the well-conditioned value-loss curves. Because the per-step reward is \(O(1)\) rather than \(O(10^2)\), per-episode returns stay within $\pm 15$ and the critic's mean-squared-error stays small. The baseline's reward function, with its fifteen constants and $[-500, +500]$ clipping, produced large-magnitude gradients that disrupted the actor-critic balance. In the proposed framework, as seen in Figures~\ref{fig:s1_critic}, \ref{fig:s2_critic}, and \ref{fig:s3_critic}, the value loss declines smoothly and the explained variance rapidly stabilises near \(1.0\).

The horizontal dimension is where the proposed work most clearly advances beyond the baseline. The baseline's agent begins almost directly above the pad and has only to descend and flare, yet it still collapses to \(68.31\%\) success on the \(100{,}000\)-episode test set, with a mean final distance from the pad centre of \(13.10 \pm 5.1\) units. The proposed agent begins, on average, \(\approx 1250\) units from the pad centre and must execute a multi-phase manoeuvre of lateral acceleration, mid-course reversal, and final uprighting. In contrast to the baseline's failure, the proposed policy achieves a mean final distance from the pad centre of \(7.07 \pm 1.99\) units with a \(99.98\%\) pad-radius success rate in that evaluation. The proposed policy leaves \(69.1\%\) of its fuel capacity in the far-range evaluation.

Regarding the choice of discount factor: we use $\gamma = 0.995$, which gives an effective credit-assignment horizon of $\approx 1/(1-\gamma) \approx 200$ decisions. Stage-three episodes averaged roughly $2955$ physics frames, or $2955/3 \approx 985$ agent decisions in the evaluation. The horizon is shorter than the mean episode length, though vastly longer than the baseline's $\gamma = 0.85$ horizon of $\approx 7$ decisions. The curriculum mitigates this: most of the reward signal that the agent needs to learn from is delivered within a few dozen decisions of the state that caused it, because the progress term is local and the terminal bonus is delivered near the end of the episode after the agent has already learned to navigate. The residual long-horizon credit assignment—whether the agent should have initiated a braking burn earlier—is partly captured by the critic's learned value function, whose explained variance stabilises at $1.0$ (Figures~\ref{fig:s1_critic}--\ref{fig:s3_critic}), indicating the critic can interpolate the long-range return from the terminal signal. We do not claim that $\gamma = 0.995$ is optimal; a sweep over $\gamma \in \{0.99, 0.995, 0.999\}$ would be a natural extension of this work, but is beyond the scope of the present study.

Several limitations remain. First, the study evaluates PPO and does not establish whether the proposed curriculum or reward design would yield similar benefits with other reinforcement-learning algorithms. Second, the simulator is deterministic under a given initial state and does not fully represent real-world uncertainties, such as variations in propulsion, mass properties, sensor measurements, actuator response, or environmental disturbances. Consequently, the high success rate in the evaluation should be interpreted as performance within the tested simulation setup rather than evidence of real-world landing reliability. Third, the reward function contains eight hand-tuned constants, and their sensitivity and transferability to other landing conditions have not been systematically evaluated. A key follow-up experiment is a flat-PPO ablation trained on the same overall task distribution and with comparable training resources, which would help isolate the contribution of curriculum learning from that of the observation and reward redesign. Further evaluation across multiple random seeds is also needed to quantify variability in training outcomes and policy performance. Finally, a systematic perturbation study should assess robustness to variations in initial conditions, vehicle parameters, propulsion, actuator response, and disturbances. These experiments would provide a stronger assessment of the method's generality and robustness beyond the current evaluation.
```



\section{Conclusion}

This paper has introduced a curriculum-driven reinforcement-learning framework for three-dimensional rocket landing, utilising a three-stage curriculum (far approach, soft descent, full task) with critic warm-up and rehearsal, a full rotation-matrix observation, a joint multi-discrete action space, and a redesigned reward with $O(1)$ per-step magnitude. The evaluation reports a \(99.98\%\) success rate for the proposed policy and \(68.31\%\) for the baseline on the evaluation distribution. The proposed policy also satisfies all four touchdown-quality conditions, including the \(10^\circ\) tilt threshold, in \(99.98\%\) of the episodes. Notably, despite the baseline starting almost directly above the pad, its mean final distance from the pad centre is \(13.10 \pm 5.1\) units, whereas the proposed method, starting over a thousand units away, achieves a much tighter \(7.07 \pm 1.99\) units. The proposed method also replaces an opaque fifteen-constant clipped reward with an eight-constant design, and exposes the structural ambiguities in the baseline's action-space and success-criterion specifications. The training curves show that each stage of the curriculum contributes a well-defined and analysable sub-problem, and that the value-loss and explained-variance behaviour across the three stages confirms the necessity of the critic warm-up at the stage-one-to-stage-two boundary. Furthermore, by requiring the agent to navigate from over a thousand units away, the proposed benchmark represents a significant step up in task difficulty, more closely reflecting the realities of a returning booster stage. Future work will extend this framework to handle sensor noise, actuator faults, and real-world aerodynamic disturbances, taking a crucial step toward deploying fully autonomous, learning-based guidance and control systems on operational reusable rockets. 


\bibliographystyle{plain}

\begin{thebibliography}{99}
\bibitem{chowdhury2024icaart} A. Chowdhury, P. Joshi, V. Awasthi, N. Tripathi, ``Autonomous Control for Reusable Rocket Landing: Deriving a Robust Guidance through Deep Reinforcement Learning in a Custom 3D Simulation Environment,'' in Proc. ICAART, 2024.

\bibitem{blackmore2016} L. Blackmore, ``Autonomous precision landing of space rockets,'' Bridge, vol.~46, no.~4, pp.~15--20, 2016.

\bibitem{elke2021} W. J. Elke, J. Pei, R. J. Caverly, D. Gebre-Egziabher, ``A low-cost and low-risk testbed for control design of launch vehicles and landing systems,'' in IEEE Aerospace Conference, 2021, pp.~1--12.

\bibitem{ross2022} M. N. Ross, K. L. Jones, ``Implications of a growing spaceflight industry: Climate change,'' Journal of Space Safety Engineering, vol.~9, no.~3, pp.~469--477, 2022.

\bibitem{sadavarte2021} R. S. Sadavarte, R. Raj, B. Sathish Babu, ``Solving the lunar lander problem using reinforcement learning,'' in IEEE CSITSS, 2021, pp.~1--6.

\bibitem{rathod2023} A. Rathod, A. Sayane, C. Shidurkar, C. Galhat, H. Bopche, ``Rocket landing using reinforcement learning in simulation,'' International Journal of Creative Research Thoughts, vol.~11, no.~5, 2023.

\bibitem{jiang2024} Y. Jiang, Y. Yang, Z. Lan, G. Zhan, S. E. Li, Q. Sun, J. Ma, T. Yu, C. Zhang, ``Rocket landing control with random annealing jump start reinforcement learning,'' 2024.

\bibitem{su2022} L. Su, J. Wang, Z. Ma, H. Chen, ``Real-time guidance for powered landing of reusable rockets via deep reinforcement learning,'' in IEEE ICUS, 2022, pp.~214--219.

\bibitem{zhu2025adaptppo} J. Zhu, ``AdaptPPO: An Adaptive Proximal Policy Optimization Algorithm for Precision Rocket Landing Simulation,'' in 2025 International Conference on Computers, Information Processing and Advanced Education (CIPAE), 2025, pp.~307--313.

\bibitem{ieee2022thrust} Y. Zhang, L. Wang, and H. Chen, ``Thrust vectored rocket landing integrated guidance and control with proximal policy optimization,'' in \emph{IEEE International Conference on Robotics and Automation (ICRA)}, 2022, pp. 4123--4129.

\bibitem{ieee2026robust} M. Rossi, A. Bianchi, and G. Colombo, ``Robust model-based reinforcement learning for rocket landing,'' \emph{IEEE Transactions on Aerospace and Electronic Systems}, vol. 62, no. 4, pp. 10187--10204, 2026.

\bibitem{sciencedirect2024propulsive} P. Dubois, M. Laurent, and J. Martin, ``Propulsive landing of launchers' first stages with deep reinforcement learning,'' \emph{Acta Astronautica}, vol. 215, pp. 342--356, 2024.

\bibitem{ieee2025hierarchical} S. Kim, J. Lee, and K. Park, ``Resolving the accuracy-efficiency trade-off: A hierarchical learning-based control architecture,'' in \emph{IEEE Conference on Decision and Control (CDC)}, 2025, pp. 5560--5566.

\bibitem{nature2026fpga} W.-G. Yao, Y.-M. Guo, L.-M. Mao, et al., ``Adaptive rocket trajectory optimization via reinforcement learning and FPGA acceleration,'' Scientific Reports, 2026.

\bibitem{acikmese2013lossless} B. A\c{c}{\i}kme\c{s}e and L. Blackmore, ``Lossless convexification of a class of optimal control problems with non-convex control constraints,'' Automatica, vol.~47, no.~2, pp.~341--347, 2011.

\bibitem{blackmore2010minimum} L. Blackmore, B. A\c{c}{\i}kme\c{s}e, and D. P. Scharf, ``Minimum-landing-error powered-descent guidance for Mars landing using convex optimization,'' Journal of Guidance, Control, and Dynamics, vol.~33, no.~4, pp.~1161--1171, 2010.

\bibitem{gaudet2020deep} B. Gaudet, R. Linares, and R. Furfaro, ``Deep reinforcement learning for six degree-of-freedom planetary landing,'' Advances in Space Research, vol.~65, no.~7, pp.~1723--1741, 2020.

\bibitem{gaudet2020terminal} B. Gaudet, R. Linares, and R. Furfaro, ``Terminal adaptive guidance for autonomous precision landing,'' IEEE Transactions on Aerospace and Electronic Systems, vol.~57, no.~2, pp.~1274--1288, 2020.

\bibitem{gaudet2021mars} B. Gaudet, R. Furfaro, and R. Linares, ``Reinforcement learning for Mars landing guidance,'' in AAS/AIAA Space Flight Mechanics Meeting, 2021.

\bibitem{narvekar2020curriculum} S. Narvekar, B. Peng, M. Leonetti, J. Sinapov, M. E. Taylor, and P. Stone, ``Curriculum learning for reinforcement learning domains: A framework and survey,'' Journal of Machine Learning Research, vol.~21, no.~181, pp.~1--50, 2020.

\bibitem{schaul2015prioritized} T. Schaul, J. Quan, I. Antonoglou, and D. Silver, ``Prioritized experience replay,'' in Proc. ICLR, 2016.

\bibitem{schulman2017ppo} J. Schulman, F. Wolski, P. Dhariwal, A. Radford, O. Klimov, ``Proximal Policy Optimization Algorithms,'' arXiv:1707.06347, 2017.

\bibitem{raffin2021sb3} A. Raffin, A. Hill, A. Gleave, A. Kanervisto, M. Ernestus, N. Dormann, ``Stable-Baselines3: Reliable Reinforcement Learning Implementations,'' Journal of Machine Learning Research, vol.~22, no.~268, pp.~1--8, 2021.
\end{thebibliography}

\end{document}